# Human-centered mechanism design with Democratic AI

**Authors:** Raphael Koster[1]†, Jan Balaguer[1]†, Andrea Tacchetti[1], Ari Weinstein[1], Tina Zhu[1], Oliver Hauser[2], Duncan Williams[1], Lucy Campbell-Gillingham[1], Phoebe Thacker[1], Matthew Botvinick[1,3] and Christopher Summerfield[1,4]

**Affiliations:**

[1] Deepmind, London, UK

[2] Department of Economics and Institute for Data Science and Artificial Intelligence, University of Exeter, Exeter, UK

[3] Gatsby Computational Neuroscience Unit, University College London, London, UK

[4] Department of Experimental Psychology, University of Oxford, Oxford, UK

*Corresponding author. Email: csummerfield@deepmind.com

† These authors contributed equally to this work

**Abstract:** Building artificial intelligence (AI) that aligns with human values is an unsolved problem. Here, we developed a human-in-the-loop research pipeline called *Democratic AI*, in which reinforcement learning is used to design a social mechanism that humans prefer by majority. A large group of humans played an online investment game that involved deciding whether to keep a monetary endowment or to share it with others for collective benefit. Shared revenue was returned to players under two different redistribution mechanisms, one designed by the AI and the other by humans. The AI discovered a mechanism that redressed initial wealth imbalance, sanctioned free riders, and successfully won the majority vote. By optimizing for human preferences, *Democratic AI* may be a promising method for value-aligned policy innovation.

**Significance Statement:** When humans act collectively to generate wealth, a mechanism is needed to share the proceeds fairly but without dampening prosperity. This problem of distributive justice has perplexed philosophers and politicians since antiquity. Here, developed a new research method in which AI systems were trained to design a redistribution scheme that players would vote for by majority. When we tested the mechanism using an economic game in which human players could earn real money, we found that it was more popular than rival mechanisms designed by human players, as well as a range of canonical baselines. Our methods, which we call "Democratic AI", may be useful for designing social and economic policies that are preferred by most humans.

## Introduction

The ultimate goal of AI research is to build technologies that benefit humans – from assisting us with quotidian tasks to addressing grand existential challenges facing society[1]. Machine learning systems have already solved major problems in biomedicine[2], and helped address humanitarian and environmental challenges[3,4]. However, an underexplored frontier is the deployment of AI to help humans design fair and prosperous societies[5]. In economics and game theory, the field known as *mechanism design* studies how to optimally control the flow of wealth, information, or power among incentivized actors to meet a desired objective, for example by regulating markets, setting taxes, or aggregating electoral votes[6,7]. Here, we asked whether a deep reinforcement learning (RL) agent could be used to design an economic mechanism that was measurably preferred among groups of incentivized humans.

The challenge of building AI systems whose behavior is preferred by humans is called the problem of "value alignment". One key hurdle for value alignment is that human society admits a plurality of views, making it unclear to whose preferences AI should align[8]. For example, political scientists and economists are often at loggerheads over which mechanisms will make our societies function most fairly or efficiently. In AI research, there is a growing realization that to build human-compatible systems, we need new research methods in which humans and agents interact[9–13], and an increased effort to learn values directly from humans to build value-aligned AI[14]. Capitalizing on this idea, here we combined modern deep RL with an age-old technology for arbitrating among conflicting views – majoritarian democracy among human voters – to develop a human-centered research pipeline for value-aligned AI research. Instead of imbuing our agents with purportedly human values *a priori*, and thus potentially biasing systems towards the preferences of AI researchers, we train them to maximize a democratic objective: to design policies that humans prefer and thus will vote to implement in a majoritarian election. We call our approach, which extends recent related participatory approaches[11,14,15], "Democratic AI".

As a first rigorous test, we deploy Democratic AI to address a question that has defined the major axes of political agreement and division in modern times: when people act collectively to generate wealth, how should the proceeds be distributed?[16–21] We asked a large group of humans to play an incentive-compatible online investment game that involved repeated decisions about whether to keep a monetary endowment or to share it with other players for potential collective benefit. We trained a deep RL agent to design a redistribution mechanism which shared funds back to players under both wealth equality and inequality. The mechanism it produced was ultimately preferred by the players in a majoritarian election.

## Results

We tested Democratic AI using a mechanism design problem based on an economic game. The game generalizes the linear public goods problem that has been extensively used to study human collective action[22,23] (**Fig. 1a**). In each of 10 rounds, each player $i$ contributes an integer number $c_i$ of coins to a public investment fund, drawing upon an endowment $e_i$ with the residual sum $e_i - c_i$ remaining in a private account (endowments may vary across players, with one player receiving more than the others). Aggregated contributions over $k = 4$ players are scaled by a growth factor $r = 1.6$ (positive return on investment; this is equivalent to a marginal per capita return [MPCR] of 0.4). The public fund is paid back to players under a redistribution mechanism which specifies the fraction of total public investment that is returned to each player, conditional on their contribution and endowment. This game admits a continuum of mechanisms for redistribution popularly associated with opposing ends of the political spectrum[19], in which returns variously depend on the contributions of self and others[23].

We illustrate the richness of the mechanism design problem in **Exp.1,** in which we measured human contributions made under three canonical redistribution principles: *strict egalitarian*, *libertarian*, and *liberal egalitarian*. Players ($n$ = 756) were assigned to groups of 4 players, with one *head* player who received 10 coins endowment and three *tail* players who received either 2, 4 or 10 coins (head/tail labels were nominal in the latter condition). Thus, endowments were *unequal* when tail players received less than 10 coins and *equal* when all players received 10 coins. Each group played multiple games each of 10 rounds, receiving the same endowment on each occasion, but experiencing each game under a different redistribution mechanism (see Supplementary Methods). Each redistribution mechanism determined the payout $y_i$ received by player $i$ as a different function of the public contribution of both self and others.

The *strict egalitarian* redistribution mechanism divides public funds equally among all players irrespective of their contributions[24]. It thus recreates the linear public goods game that, for $r <$

$k$, is a social dilemma in that each individual benefits from withholding contributions and free riding on the largesse of other players[23]. Accordingly, contributions under this mechanism decline over time (effect of time on contributions, $p < 0.001$; **Fig. 1c,** top panels), mirroring previously described results[22,23]. The *libertarian* mechanism[21] returns a payout to each player in proportion to their contribution $y_i = r \cdot c_i$ such that $c_i = e_i$ is a pareto-efficient Nash equilibrium. This mechanism effectively privatizes contributions and removed the social dilemma, which encouraged players to increase their contributions ($p < 0.001$; **Fig. 1c**, middle panels) as observed previously[25] (note that whilst players receive detailed instructions about the game dynamics, they are obliged to learn about each mechanism from experience). Finally, *liberal egalitarianism* proposes that each player is accountable for their actions but not initial advantage, and so payout depends on the fraction of endowment that is contributed[26]. When payouts were relative to endowment-normalized contributions (*liberal egalitarian*), the tail players learned rapidly to contribute ($p < 0.001$) but the head player's contributions remained flat ($p > 0.15$), diminishing the availability of public funds.

Previous reports have suggested that heterogeneity of endowment or MPCR can influence contribution to the public fund[27,28], especially when inequality is made salient to participants[29]. Here, we observed a comparable phenomenon when we examined the contributions of the head player (who received 10 coins) as a function of the endowment received by tail players, which could be either equal or lower. Under *strict egalitarian* the head player contributed the same irrespective of the endowment of others ($p = 0.74$), but under *libertarian* and *liberal egalitarian* the head player was less prone to contribute when others were less well off ($F_{2,91} = 8.94$, $p < 0.001$ and $F_{2,96} = 18.83$, $p < 0.001$ respectively). Thus, productivity was dampened under conditions of greater inequality.

More generally, Exp.1 highlights the challenge that the game poses for the mechanism designer: a redistribution scheme might be unpopular because it provokes a general collapse of contributions due to free riding, because it leads to unequal outcomes, or because it siphons funds away too aggressively from the wealthiest player, who then fails to provision the public fund. We thus asked whether an AI system could design a mechanism that humans preferred over these alternatives.

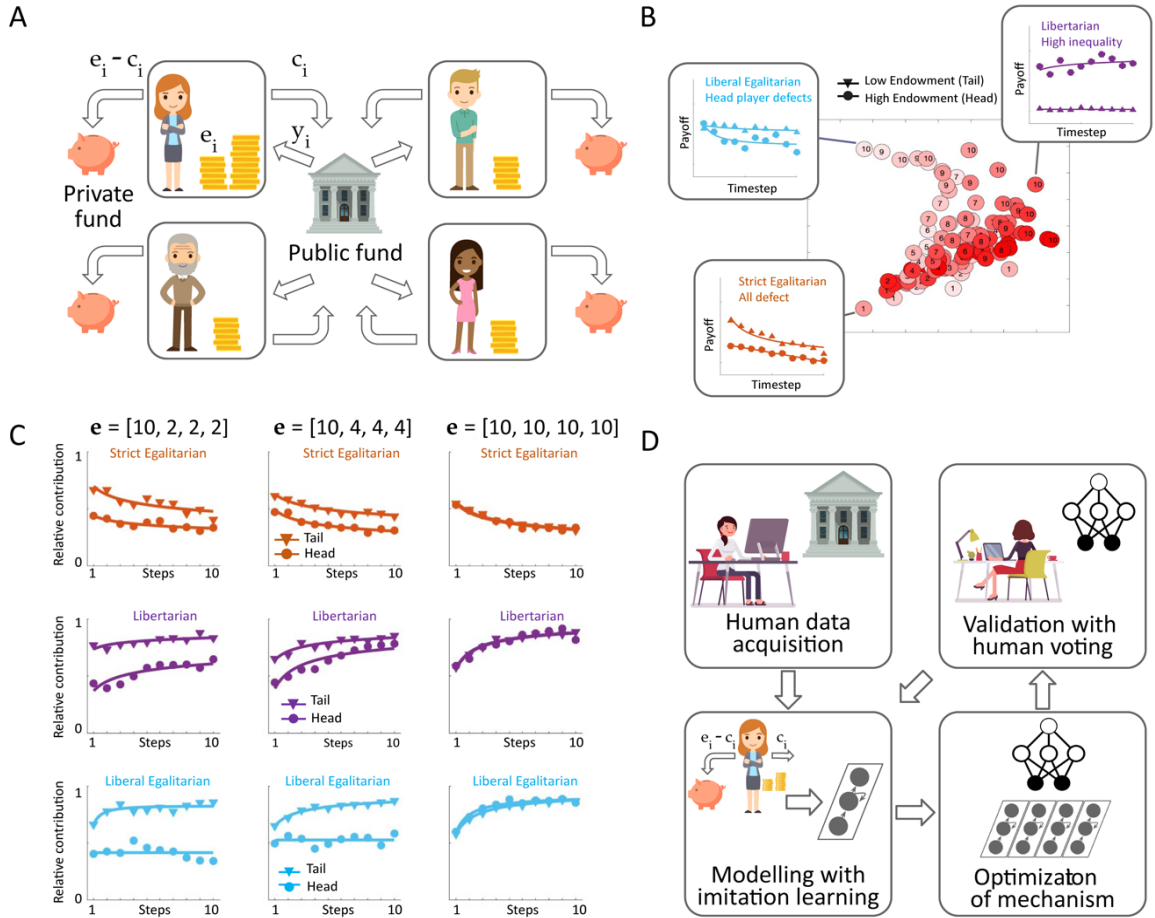

**Figure 1. A.** Illustration of the setup of the investment game. **B.** The *ideological manifold* for endowment distribution [10, 2, 2, 2]. The plot shows a visualization of a space of redistribution mechanisms defined by parameters $w$ and $v$ in two dimensions. Each red dot is a mechanism, and distances between dots conserve dissimilarities in the (average) relative payout to virtual players (both head and tail). Dot numbers denote bins of mechanism parameter $w$ (1 = lowest, 10 = highest) and shading denotes bins of $v$ (light = more relative, dark = more absolute). Inset, examples payouts to head (circles) and tail (triangles) player under the canonical mechanisms uses as baselines against which to test the AI. Under *strict egalitarian* payouts decline to head and tail players. Under *libertarian*, there is great inequality between head and tail players. Under *liberal egalitarian*, the head player stops contributing, so payouts decline for both head and tail player. **C.** Average relative contributions (as a fraction of endowment) over 10 rounds (x-axis) in Exp.1 for three different initial endowment conditions. Under *strict egalitarian* redistribution, tail player (triangles) contributions are higher when initial endowments are lower, but head player (circles) contributions do not differ. Under *libertarian*, head player contributions increase with equality, but tail player contributions remain constant. Tail player contributions increase strongly with endowment under *liberal egalitarian*. **D.** Illustration of our agent design pipeline.

How, then, should the public funds be shared? The maximally popular policy could be one of these three canonical mechanisms or something else entirely. The size of the potential search space makes it hard to identify the preferred mechanism using traditional behavioral research methods. We thus developed a human-in-the-loop AI research pipeline to tackle this problem (**Fig. 1d**). First, we collected an initial sample of human data (*Acquire*) and used it to train "virtual human players" which were recurrent neural networks that learned to imitate human behavior during the game and that voted according to the same principles as human players (*Model*; see Fig. S1). This simulation step was necessary because training agents during online interaction with humans would have been prohibitively costly and time-consuming. Thirdly, we optimized the mechanism design with deep RL, using a variant of policy gradient methods[30], to maximize the votes of virtual human players (*Optimize*). Fourthly, we sampled a new group of humans, and pitted the RL-designed redistribution mechanisms against rival baselines in a series of head-to-head majoritarian elections. This new human data was then used to augment our player modelling process, which in turn improved optimization and led to potentially better mechanisms (*Repeat*). This pipeline builds on recent approaches that have used human data interactively to train artificial agents[31–33]. We iterated this procedure to obtain a mechanism that we call the Human Centered Redistribution Mechanism or *HCRM*, which is the major focus of the remainder of this report.

Before evaluating *HCRM* with a new group of human players, we used our research pipeline to determine which baseline mechanisms might pose the strongest competition. To achieve this, we generalized the three canonical baselines to produce a continuously parameterized space of redistribution mechanisms. We first assume that the fractional payout to each player $y_i$ is composed of an absolute ($y_i^{abs}$) and relative ($y_i^{rel}$) component, which are combined via a mixing parameter $v$. These components are in turn are given by contributions from both the focal and other players (mixing parameter $w$; see below and *Methods*). In these baselines, the payout to player $i$ is thus given by

$$y_i = v\big(y_i^{rel}\big) + (1 - v)(y_i^{abs})$$

where the absolute component combines their own contribution $c_i$ with the average of that from other players $c_{-i}$ so that

$$y_i^{abs} = r[w(c_i) + (1 - w)(c_{-i})]$$

and the relative component is similarly determined by $\rho_i = c_i/e_i$ which is the ratio of contribution to endowment for player $i$:

$$y_i^{rel} = r\left(\frac{C}{P}\right)[w(\rho_i) + (1-w)(\rho_{-i})]$$

where $c_{-i}$ and $\rho_{-i}$ are respectively the average contributions and ratios from players other than $i$, and $C$ and $P$ are the sum of contributions and ratios across all players.

We call this space of baseline mechanisms defined by $v$ and $w$ the *ideological manifold* (**Fig. 1b**) We note that the three baseline mechanisms we have considered so far lie within this space: *libertarian* ($w = 1, v = 0$), *liberal egalitarian* ($w = 1, v = 1$) and *strict egalitarian* ($w = 1/k$).

We sampled mechanisms from the ideological manifold and pitted them against one another exhaustively in a two-player tournament. Our goal was to identify that which maximized votes among the virtual human players (neural networks that had been trained to imitate real human behavior). This exercise identified *liberal egalitarian* as the single Nash equilibrium of the two-player tournament and thus not only as the strongest competitor among the three canonical baselines but among the entire ideological manifold (Table S2).

Armed with this intuition, in **Exp. 2a-c** we evaluated the AI-designed *HCRM* against the three canonical baselines introduced above. Groups of four human participants ($n = 2508$) played successive incentive-compatible games of 10 rounds under two rival mechanisms, before voting for one that they preferred to play again (for additional payoff) in a final round. We randomized players into five endowment conditions, in which a "head" player received $e_{head} = 10$ coins endowment and three "tail" players received the same $e_{tail} \in \{2, 4, 6, 8, 10\}$ (once assigned, endowments remained constant throughout the game). We found that *HCRM* was more popular than all three baselines (**Fig. 2a-c** and Table S1a), obtaining a total of 513/776 (66.2%) votes against *strict egalitarian* (p < 0.0001), 450/740 (60.8%) against *libertarian* (p < 0.0001) and 951/1744 (54.5%) against *liberal egalitarian* (p < 0.002).

Against *strict egalitarian* and *libertarian,* the AI-designed mechanism was also more popular under all five endowment distributions tested, ranging from full equality to the most unequal endowment condition ([10, 2, 2, 2] implies a Gini coefficient of 0.38, roughly equivalent to contemporary Russia). Across these conditions, its vote share ranged from 56.0% to 67.0% against *egalitarian* and from 57.5% to 66.7% against *libertarian*. Consistent with the data from our 2-player tournament, liberal egalitarian proved popular among humans and thus more difficult to beat. Indeed, although *HCRM* was preferred overall, under full equality (64.5%, p < 0.001) and under moderate inequality (endowments [10, 8, 8, 8] and [10, 6, 6, 6]) with a vote share of 54.5% (p < 0.02), there was no reliable difference in voting preference between *HCRM* and *liberal egalitarian* under the most unequal conditions (*HCRM* vote share 47.4%, p = 0.19), suggesting that *liberal egalitarian* redistribution offered an equally good alternative to *HCRM* under conditions of highest inequality.

Our AI-designed *HCRM* was trained by interacting with neural networks that imitated human behavior. However, if our participants are rational agents who learn to maximize their return over the course of each game, then it should be possible to solve the problem without recourse to human training data at all, by substituting our virtual human players for a new class of *rational players* that are trained to maximize their own expected return within the game (see *Methods*). Previous work has implied that successful human-centered mechanisms can be obtained in this fully multi-agent setting[34,35]. Alternatively, if modelling human cognitive biases is critical then a system trained to maximize the votes of *rational players* may transfer more poorly back to human participants. In **Exp.3**, we tested this by exposing a new group of human participants ($n = 736$) to both the mechanism designed by *HCRM* and that proposed by a new *rational mechanism* or *RM* that was trained with *rational players* but otherwise identical.

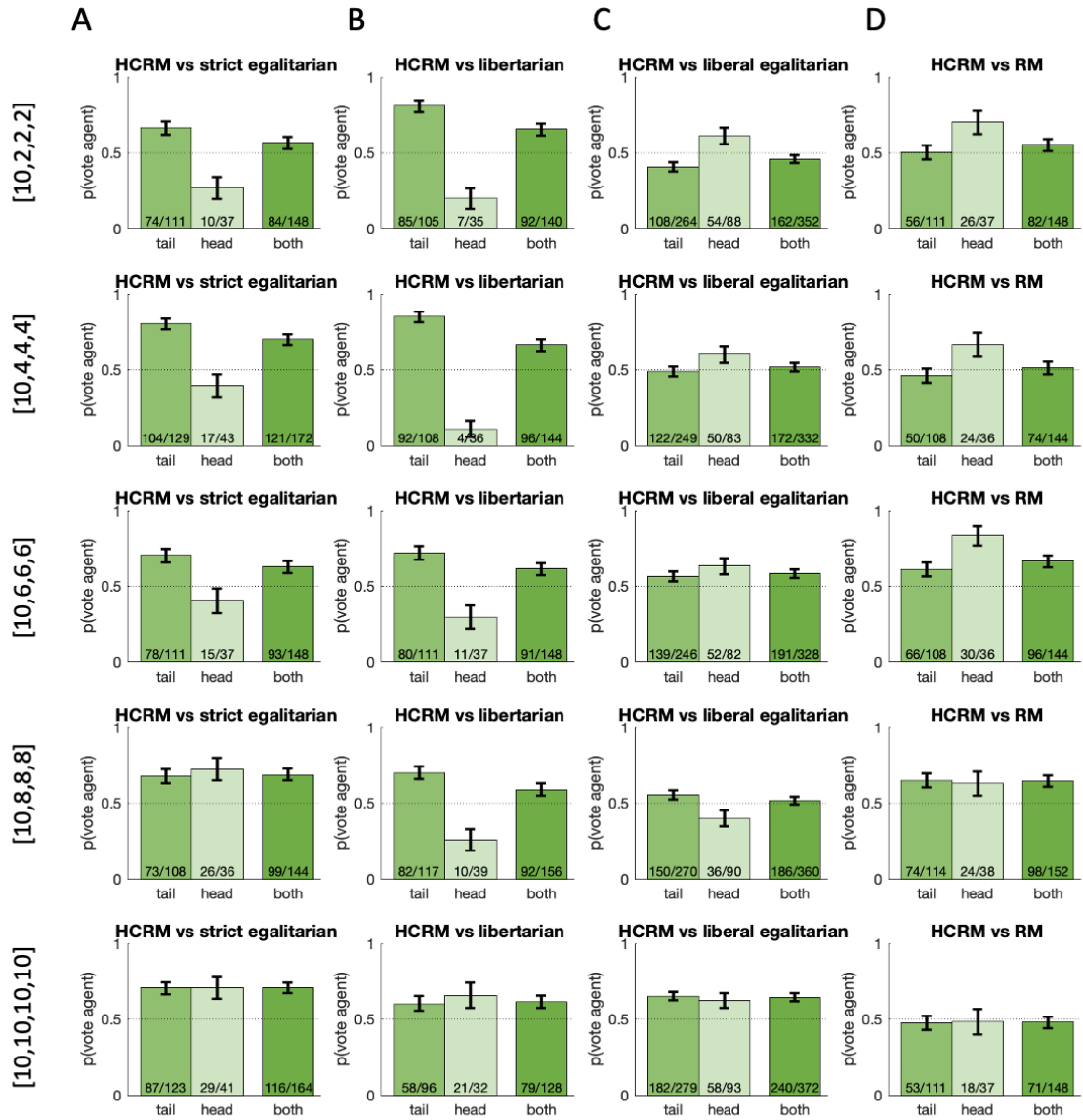

**Figure 2.** Vote share for the *HCRM* against the three canonical baselines (leftmost columns) and the *Rational Mechanism* for each endowment condition (rows). The three bars show the average number of votes for the agent given by the tail players, the head player, and all players.

Overall, 57.2% of participants preferred *HCRM* over the *RM* (p < 0.0003; **Fig. 2d**). Interestingly, *RM* learned a radical policy under unequal endowments that neglected the head player and paid out principally to the tail players (**Fig. S3**). Despite the favorable ratio of tail to head players, however, this was unsuccessful even in the most unequal endowment conditions, because the head player rapidly stopped contributing to the detriment of everyone (including the tail players) leading to an overall lower group surplus than for *HCRM* ($t_{183} =$ 7.96, p < 0.001). In other words, the redistribution policy that humans prefer is neither one that shares out public funds equally, nor one that tries to speak only to the interests of a majority of less well-endowed players. One exception was under equal endowment, where *HCRM* and *RM* performed nearly identically (*HCRM* vote share 47.9%, p = 0.49), implying that for the setting we explored, rational models may offer a good account of human behavior when the initial conditions are fair. Together, however, these results imply that human data may be crucial when using AI for mechanism design.

The RL mechanism designer was not equipped with memory, and so *HCRM* is readily interpretable – it can be transparently described as a two-dimensional surface that maps the relative contribution of head and tail players to their share of the proceeds ('Beach plots' in **Fig. 3A**). This allowed us to ask why the RL-designed mechanism is popular with human players. RL discovered a hybrid mechanism that eschewed traditionally proposed redistribution schemes that emphasize individual discretion over resource allocation (*libertarian*) or collective equality (*strict egalitarian*). Pursuing a broadly *liberal egalitarian* policy, *HRCM* sought to reduce pre-existing income disparities by compensating players in proportion to their contribution *relative to endowment*. In other words, rather than simply maximizing efficiency, the mechanism was progressive: it promoted enfranchisement of those who began the game at a wealth disadvantage, at the expense of those with higher initial endowment. In doing so, it achieved a favorable trade-off between productivity (surplus) and equality (Gini coefficient) among rival mechanisms (**Fig. 3b;** see also Fig. S3). However, unlike *liberal egalitarian*, it returned almost nothing to players unless they contribute approximately half their endowment (**Fig. 3c**). In other words, RL effectively discovers that humans facing social dilemmas prefer mechanisms that allow for sanction of free riders[36]. The agent thus learns a policy that is not readily assigned to a specific philosophy of distributive justice but creatively combines ideas from across the political spectrum.

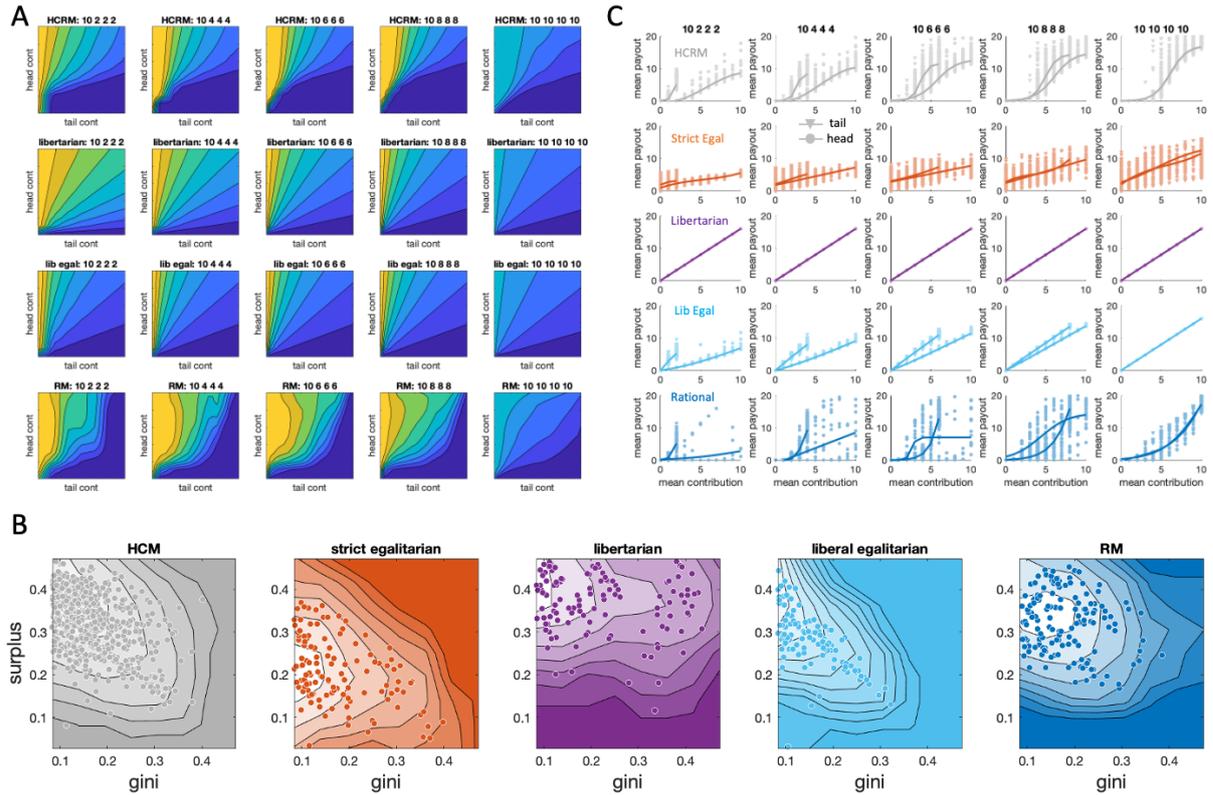

**Figure 3. A.** 'Beach plots': In simulation with virtual players, revenue share allocated to head player as a function of contribution (relative to endowment) of head vs. tail player. Warmer (colder) colors indicate more funds redistributed to head (tail) player. **B.** Two-dimensional distributions of surplus (in log units) and Gini Coefficient for each group under each mechanism. Lighter colors indicate higher density. Higher surplus implies greater productivity; lower Gini implies greater equality**.** Each dot is a game. **C.** Empirically observed relationship between contributions and payouts for each mechanism and endowment condition. Each dot is a head player or the average of tail players in a single game. Lines are fit separately to head and tail players.

Finally, we asked whether trained and incentivized human players could have devised a mechanism that was as popular as *HCRM*. We first recruited 61 previous players and trained them, over the course of about an hour, to redistribute funds to virtual citizens with a view to maximizing votes (i.e., the same training regime as our agent). Human referees earned £2 per vote. Focusing on the [10, 4, 4, 4] condition, which was among those that *HCRM* found most challenging, we then recruited an additional set of new human players (n = 244), who played one game under *HCRM* and another with trained human referees, in counterbalanced order. These human players strongly preferred *HCRM* over the human referee (63.3% voted for *HCRM*, p < 0.0001). Interestingly, the human players were overall less prone to sanction the head player with low payouts (government x payout sextile interaction, $F_{2,128} = 5.541$, $p < 0.005$; **Fig. 4B**) and failed to reward the tail players sufficiently for contributing generously from the little they had (**Fig 4C,** lower panel) relative to HCRM (upper panel). We show empirically derived beach plots for the human and algorithmic referees side by side in **Fig. 4D**; they imply that overall, human referees were less responsive to contributions when allocating payouts.

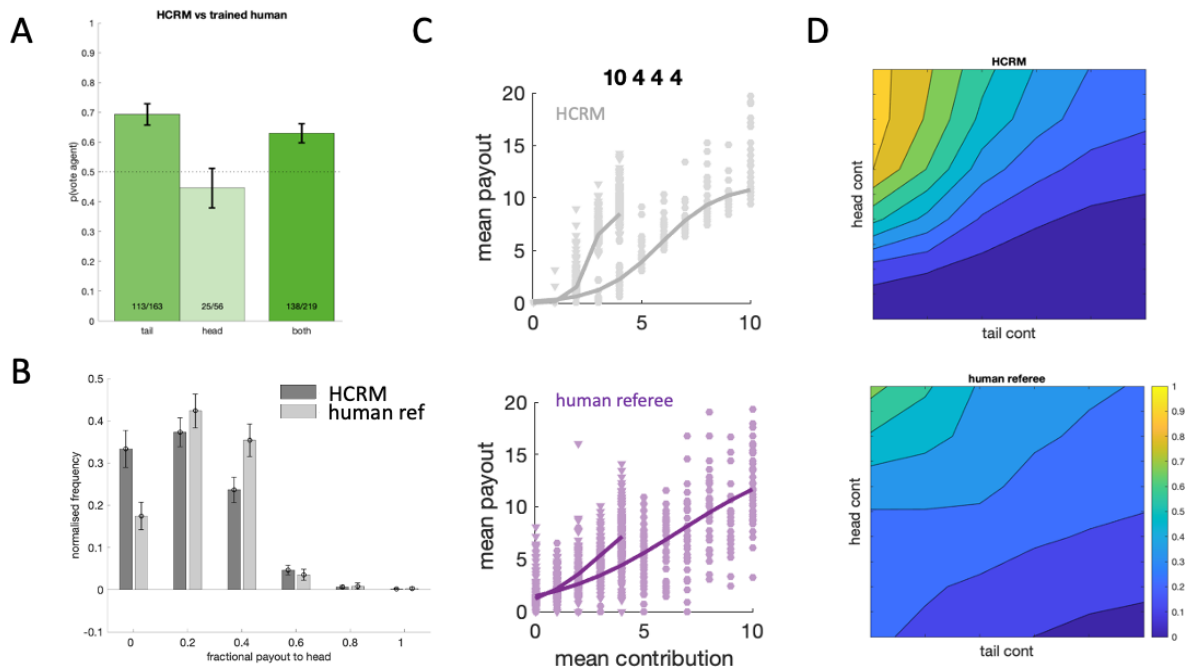

**Figure 4. A.** Votes for the agent against trained humans. **B.** Normalized frequencies of fractional payouts to the head player in 6 equally spaced bins. Note the HCRM is more willing to offer low payout to head player than human referees (leftmost bar). **C.** Empirically observed relationship between contributions and payouts for each HCRM (upper panel) and the human referee (lower panel). Each dot is a head player (circle) or average of tail players (triangle) in a single game. The slope mapping contributions onto payouts for the tail player is shallower for human referees. **D.** Empirically observed relationship between payout and head/tail player contributions ("beach plot") for *HCRM* (top) and human referee (bottom) for [10, 4, 4, 4]. The color scale shows the fraction of the public fund allocated to the head player, as a function of the head (y-axis) and tail (x-axis) player relative contributions.

## Discussion

Together, thus, these results demonstrate that an AI system can be trained to satisfy a democratic objective, by designing a mechanism that humans demonstrably prefer in an incentive-compatible economic game. Our approach to value alignment relieves AI researchers – who may themselves be biased or unrepresentative of the wider population – of the burden of choosing a domain-specific objective for optimization. Instead, we show that it is possible to harness for value alignment the same democratic tools for achieving consensus that are used in wider human society to elect representatives, decide public policy, or make legal judgments.

Our research raises several questions, some of them theoretically challenging. One might ask whether it is a good idea to emphasize a democratic objective as a method for value alignment. Democratic AI potentially inherits from other democratic approaches a tendency to enfranchise the many at the expense of the few: the "tyranny of the majority"[37]. This is particularly pertinent given the pressing concern that AI might be deployed in way that exacerbate existing patterns of bias, discrimination or unfairness in society[38]. In our investment game, we sampled endowment conditions to match plausible real-world income distributions, where the disadvantaged inevitably outnumber the advantaged – and so for the specific question of distributive justice that we address, this problem is less acute. However, we acknowledge that if deployed as a general method, without further innovation, there does exist the possibility that (just like in real-world democratic systems) it could be used in a way that favors the preferences of a majority over a minority group. One potential solution would be to augment the cost function in ways that redress this issue, much as protections for minorities are often enshrined in law. We offer a more extensive discussion of these issues in the Supplementary Text.

Our AI system designed a mechanism for redistribution that was more popular than that implemented by human players. This is especially interesting because unlike our agent, human referees could integrate information over multiple timesteps to reward or sanction players based on their past behaviour. However, on average the human-invented redistribution policy tended to reward the tail player insufficiently for making relatively large contributions (from their smaller endowment) to the public purse, was less popular than that discovered by HCRM. Humans received lower volumes of training data than HCRM, but presumably enjoyed a lifetime of experience with social situations that involved fair and unfair distribution, so we think they represent a strong baseline, and a proof of concept for AI mechanism design.

Finally, we emphasize that our results do not imply support for a form of "AI government", whereby autonomous agents make policy decisions without human intervention[39,40]. We see *Democratic AI* as a research methodology for designing potentially beneficial mechanisms, not a recipe for deploying AI in the public sphere. This follows a tradition in the study of technocratic political apparatus that distinguishes between policy development and policy implementation, with the latter remaining in the hands of elected (human) representatives[41]. We hope that further development of the method will furnish tools helpful for addressing real-world problems in a truly human-aligned fashion.

**Acknowledgements.** We thank Michiel Bakker, Martin Chadwick, Hannah Sheahan, Koray Kuvakcuoglu, Demis Hassabis and Laura Weidinger for comments on an earlier version of this paper. We thank Christiane Ahlheim, Josh Merel, Doug Fritz, and Miteyan Patel for technical assistance and/or helpful suggestions.

**Funding.** None to declare.

**Author contributions.** Conceptualization, Formal Analysis, Investigation, and Writing: RK, JB, AW, AT, CS, MB, OH, TZ. Project Administration: LC, PT. Supervision: CS, MB. Software and Resources: DW.

**Competing interests:** None to declare.

**Data availability:** All data and relevant code will be made available on publication.

**Supplementary Materials**
Materials and Methods
Text S1 – S3
Fig S1 – S9.
Table S1 – S2.

# Materials and Methods

## Participants

Participants were recruited over an approximately 8-month period from the crowdsourcing platforms Prolific Academic and Amazon Mechanical Turk. All participants gave informed consent to participate in the experiment. The task was advertised to users located in the UK and USA. We did not record identifiers or personal information from participants. Participants who accepted the Human Intelligence Task (HIT) received a link that led them to a game lobby in which they were grouped with three other players. Groups of four players participated in the game live and in interaction with one another. When a response was required, participants had a fixed duration in which to respond (2 mins for all screens except voting, which was 4 mins), accompanied by a timer which signaled how long they had remaining. The game advanced only when the player who was slowest to respond had completed the round. Players who timed out were given a warning. Players who timed out twice were removed from the game and replaced with a randomly responding bot (games with missing data were excluded from the analysis). The game took approximately 20-30 minutes and participants were paid up to £8 consisting of a base and a bonus payment. The precise conversion rate of points earned in-game (return in coins) to the bonus paid at the end of the study varied inversely with the sum of endowments over players. This way stakes were on average equated across games.

A total of n = 4776 participants took part in experiments 1-3. The pilot data that was used for training the agent consisted of a further ~4000 datasets (including some partial datasets where participants timed out). Exclusion lists were used to prevent participants from rejoining the experiments multiple times (however, as the overall data was collected over several months on two platforms, and we did not collect identifiers, it was impossible to be absolutely sure that all participants are unique). The study was approved by HuBREC (Human Behavioral Research Ethics Committee), which is a research ethics committee run within Deepmind but staffed/chaired by academics from outside the company.

## Investment game

All participants in Exp.1-3 played 34 rounds of an investment game (3 blocks of 10 rounds and 1 "bonus" block of 4 rounds). On each round, each player was allocated an endowment of 2, 4, 6, 8 or 10 coins (2, 4 or 10 coins in Exp.1) depending on the endowment condition to which they were allocated, and whether they were designated the "head" or "tail" player, all of which was entirely random (and unrelated, for example, to the order in which they joined the game). In all endowment conditions there was a single head player who received 10 coins and three tail players who all received either 2, 4, 6, 8 or 10 coins (in the "equal" endowment condition the distinction between head and tail players is nominal only). Players received their endowment at the start of each round. Each players' endowment remained the same across all 34 rounds (and they were instructed that this would be the case).

In every round of every block, each player $i$ privately chose to divide an integer number of coins (their endowment $e_i$) between a "project" and a "private account" (the contributions made to the project are denoted $c_i$). No player could see the others' choices at this stage. The "project" was a public fund that received a return on investment (was multiplied by a common productivity factor $r = 1.6$) and was then shared between participants according to some redistribution scheme, allocating a payout $y_i$ to player $i$ (see below). Coins allocated to the "private account" were simply retained by participants (with no return on investment). The total return to each player on each round was thus their payout plus endowment minus contribution $y_i + e_i - c_i$.

In the first block of all experiments, participants played 10 "tutorial" rounds with no referee. This meant that funds allocated to the project were distributed equally among all players and there was no further redistribution (see below). In blocks 2 and 3 participants played 10 rounds with a referee (or mechanism; one mechanism for each block). The referee(s) redistributed project funds among players according to a specified mechanism, without creating or destroying wealth. The two rival mechanisms were encountered in counterbalanced order. After block 3 participants voted for the mechanism that they preferred. They knew that they would be making this vote (and what it would entail) from the end of block 1, before experiencing the mechanisms. In block 4, the probability of re-experiencing mechanism A (or B) was exactly equal to the fraction of votes that A (or B) received from the 4 players. The choice was thus deterministic if all players voted the same way, and there was no opportunity to vote strategically. Participants then answered 7 debriefing questions (see below). Finally, they experienced four rounds of the chosen mechanism (block 4) and proceeded to a bonus screen where they were thanked and informed of their total earnings. Only data from blocks 2 and 3 were included in the analysis of Exp. 2-4 (see below for Exp. 1).

## Detailed description and illustration of the game

Participants began the task with instructions and a tutorial. Players were initially told that they were playing an investment game and could "earn points depending on both [their] own decisions and on the decisions of others" and that they would "receive a base payment for completing the task but could also get a bonus, "depending on how many points [they] earn". They then underwent a demonstration round with identical structure to each of the real rounds in the game, except that their choices were not logged.

On every round each player was allocated an endowment consisting of an integer number of coins and decided to contribute each coin to a "project" or a "private account" by pressing buttons with those corresponding labels (see image below). After each response a coin disappeared from the endowment row and appeared below the button that had been pressed (project or private account, see below). When all coins had been allocated, the "submit response" button became available, and participants pressed this to complete the trial. The screens below show an example for a player whose endowment was 10 coins:

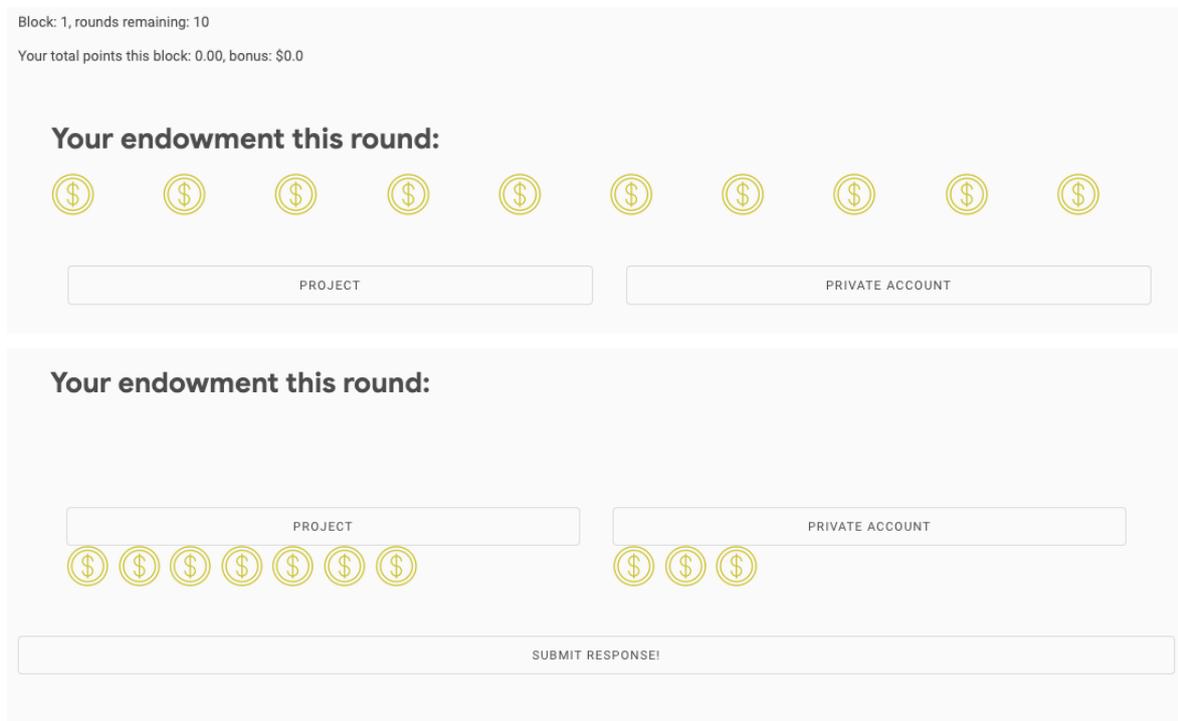

After all players had completed the round by allocating their full endowment of coins and pressing "submit response" they viewed a results screen (image below). On all rounds, three panels were shown (Fig. S2). The upper panel was a bar graph that showed contributions to the private account and project made by each of the 4 players, including the focal player, both numerically and in the form of a bar graph. Players were denoted with a distinctive icon. The bar graph was always scaled so the y-axis ranged from 0-10 coins.

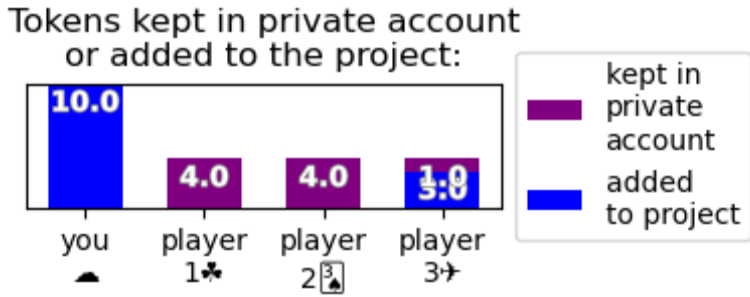

The second panel differed according to whether players were in the training game (block 1) or the test or bonus games (blocks 2-4). In the training game, there was no "referee". An example of the screen shown in this block is given below. It reveals the "project earnings" for each player (this is called "payout" in the main text and denoted $y_i$ for player $i$). In the example, 10 coins have been allocated to the project (6 from the focal player, and four from player 3). These are increased to 16 coins following the multiplication factor of $r = 1.6$. Note that when there is no referee, distribution is always equal between players, which is identical to the mechanism "strict egalitarian".

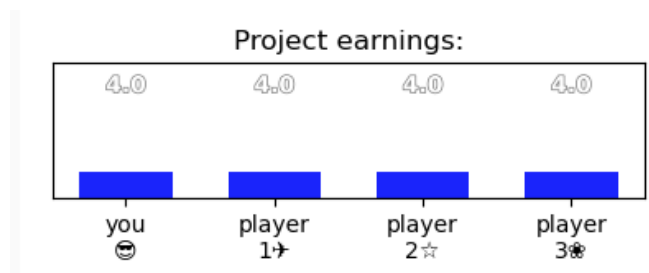

In blocks 2-4 there was a "referee", which was the AI-designed mechanism *HCRM* or a rival baseline. In either case, participants viewed a screen that looked like this:

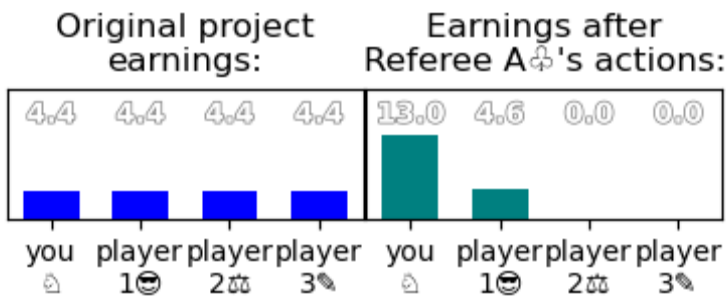

The screen on the left is the same as for block 1, except that the title "project earnings" becomes "original project earnings". The referees' actions were framed as an intervention on a default distribution, which is identical to 'strict egalitarian'. On the right hand side, players saw an additional panel that shows how these earnings have been redistributed among players by the referee, hence 'after' the referee's action. The referees were referred to as "A" and "B" and also identified by a unique color and symbol. All earnings before and after redistribution were displayed to one decimal place as in the example. The bar color for the rightmost panel varied according to the referee identity.

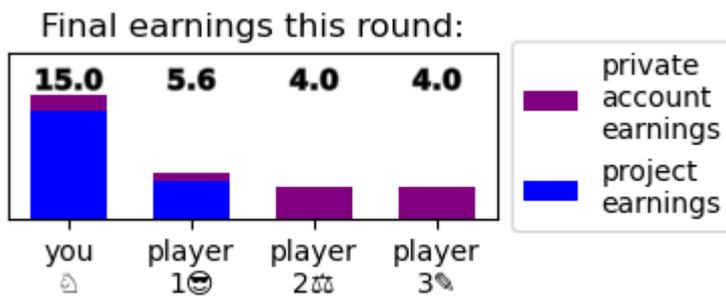

In all blocks, the final panel of the results screen gives participants a final overview about how much return was obtained by each participant, divided earnings returned from the project and by what was kept in the private account.

## Voting

An illustration of the voting screen is shown below:

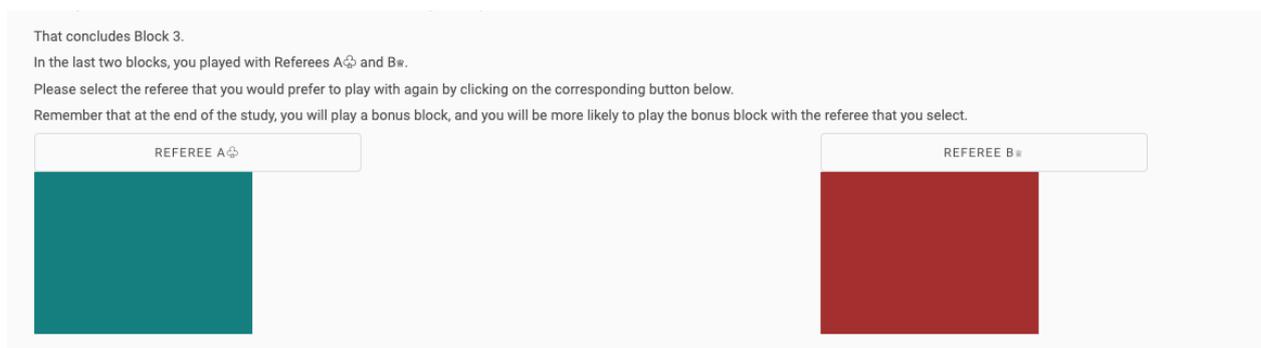

Participants were instructed to make a binary choice between the two referees (each marked with a different letter, symbol and color which matched those seen during the results screen for each mechanism). It was highlighted to participants that their choice will affect the likelihood of playing the last block with the referee they chose. After their selection, a button appeared that asked participants to confirm their choice.

# Debriefing

After the vote (but before block 4) participants answered 7 binary debriefing questions:

1. Which referee will lead to everyone doing better?
2. Which referee will lead to you individually doing better?
3. Which referee will lead to the majority of players doing better?
4. Which referee will be fairer?
5. Which referee will be best at encouraging collaboration?
6. Which referee will be more permissive or lenient?
7. Which referee will be more transparent or predictable?

Participants responded by checking a radio button next to either "referee A" or "referee B" following each question. The results of this debriefing questionnaire are shown in Fig. S6.

# Detail of Experiments

We describe 3 experiments in the main text. All experiments had the same form but the way we present the data is slightly different for Exp. 1 relative to the other two experiments Exp. 2 and Exp. 3.

In Experiment 2 (n = 2508) players experienced the *HCRM* and either *strict egalitarian, libertarian* or *liberal egalitarian* mechanisms in randomized groups (we provide details about all mechanisms below). The order of encounter of the different mechanisms was counterbalanced over blocks 2 and 3. Under the *strict egalitarian* mechanism, the referee effectively took no action, so that the original earnings and "earnings after referee's actions" screens looked the same (apart from bar color). Under *libertarian*, the mechanism returned to each player a sum that was 1.6x their contribution. We decided to recruit a larger cohort for *liberal egalitarian* because simulation data (Table S2) suggested that this was the highest performing baseline and potentially more data would be required to draw a reliable conclusion about which was preferred (target participant numbers were decided beforehand, and we did not use optional stopping). In Experiment 3 (n = 736) players experienced *HCRM* and the "rational mechanism" (RM; the rational mechanism was one trained with rational players; see below for description of these players, and main text for more details about the RM) in counterbalanced order over blocks 2 and 3.

The data described as "Experiment 1" came from a study in which an earlier version of the HCRM competed against *libertarian* and *liberal egalitarian*. We present data only from these two baselines (not the earlier version of the agent). The data from *strict egalitarian* are taken from block 1 (in which there was no referee). Thus, the conditions under which these data were collected are the same as Exp.2, except that (i) the order of the mechanisms was not counterbalanced, (ii) the rival AI-designed mechanism was slightly different, and (iii) we do not report voting data from this experiment. Our goal here is to illustrate how contributions vary under different mechanisms; in fact, a near-identical pattern of results is replicated in Exp.2. At this stage we only used three endowment conditions: [10, 2, 2, 2], [10, 4, 4, 4] and equality ([10, 10, 10, 10]).

## Data pre-processing and analysis

Our main analyses focus on data from Exp 1-3 (n = 4776 total). We have made this large dataset (from Exp.1-3) freely available at [link available on publication].

In Exp. 1 we plot contributions as a function of mechanism and endowment as described above (**Fig. 1c**). In Exp. 2-3, data were analyzed from blocks 2 and 3, that is from the two blocks of 10 trials in which participants played the game with a referee that was either the *HCRM* or a rival baseline. We logged or computed various per-block, per-round metrics for each individual player, including [absolute] contributions ($c_i$), relative contributions ($c_i/e_i$), [absolute] payouts ($y_i$), relative payouts ($y_i/e_i$), and return ($e_i - c_i + y_i$), as well as some group-level game metrics including Gini coefficient and surplus (sum of returns / sum of initial endowments over players and rounds). We show contributions and payouts over time for head and tail players under each mechanism in Fig. S3 and Fig. S4.

Our voting data consisted of 4 binary votes per group $g$ which were either for HCRM or a baseline. We performed a group-level permutation test to assess statistical significance. Our permuted data randomly flipped voting preferences but preserved the covariance among votes within a group (note that statistics from a naive binomial test might be inflated as they are not independent conditioned on participants' shared experiences of the mechanism). The lowest possible p-value obtainable given our 10K shuffles is p < 0.0001.

## Determinants of voting

We conducted an analysis to understand the determinants of voting using logistic regression. Results are shown in Fig. S2.

For votes in Exp. 2-3 we constructed a regression model with the following form:

$$p(vote^{HCRM}) = \Phi[\beta_0 + \beta_1 \cdot rpay + \beta_2 \cdot apay + \beta_3 \cdot cont]$$

Where $rpay$, $apay$, $cont$, and $gini$ are relative variables that encode an aggregate quantity (over 10 rounds) under the $HCRM$ minus that same quantity under the rival mechanism ($rpay$ is payout divided by endowment for the focal player; $apay$ is absolute payout to the focal player; and $cont$ is the sum of contributions of all group members. We tried other regressions that included Gini coefficient and/or payout to other players, but these fit less well. The logistic function is denoted $\Phi[\cdot]$. Predictors were standardised before being entered into the regression, so the y-axis in Fig. S2 shows units of standard deviation.

We report statistics for $\beta_0$ to $\beta_3$ in an additional table.

## Beach plots

In order to visualize the $HCRM$, we plotted the fraction of payouts that were allocated to the head player as a function of an exhaustive set of possible contributions (relative to endowment) made by simulated head and tail players. Note that because the $HCRM$ is time-invariant there exists only a single such surface (not one per timepoint). To aid visualization, we averaged across the contributions of the tail player, reducing a four-dimensional matrix to just two dimensions. The plot in **Fig. 3a** (main text) thus shows the average fraction of returns to head player (1-tail player) for each combination of head and (average) tail contributions. In the Fig. S8 we show equivalent plots for which the dependent measure is the return to either the head or tail player.

## The ideological manifold

We defined a space of mechanisms that spanned *strict egalitarian*, *libertarian*, and *liberal egalitarian* schemes as follows. First, we assumed that the payout to player $i$ is an admixture of their own (absolute) contribution $c_i$ and the average of other players $c_{-i}$ so that

$$y_i^{abs} = r[w \cdot c_i + (1 - w) \cdot c_{-i}]$$

where $w$ is the mixing parameter. Thus, for example, if $w = 1/k$ ($w = 0.25$ for our case of 4 players) then $y_i^{abs}$ will be equal for all players. But we can also define a space of relative mechanisms:

$$y_i^{rel} = r \left(\frac{C}{P}\right) [w \cdot \rho_i + (1 - w) \cdot \rho_{-i}]$$

where $\rho_i = c_i/e_i$ is the ratio of contribution to endowment for player $i$, where $c_{-i}$ and $\rho_{-i}$ are respectively the average contributions and ratios from players other than $i$, and $C$ and $P$ are the sum of contributions and ratios across all players. We can then define a full space of baseline mechanisms defined by both $w$ and $v$ as a mixture of absolute and relative redistribution:

$$y_i = v(y_i^{rel}) + (1-v)(y_i^{abs})$$

Thus if $w = 1$ and $v = 0$ then payout is exclusively driven by $y_i^{abs}$ and thus reduces to $y_i = r \cdot c_i$, which is the *libertarian* mechanism. Alternatively, if $w = 1$ and $v = 1$ then payout is exclusively driven by $y_i^{rel}$, and thus depends on $\rho_{-i}$ which is the ratio of contributions to endowment, which is the *liberal egalitarian* mechanism.

We visualized the space of mechanisms defined by mechanism parameters $v$ and $w$ (see below and main text for details) in terms of the relative payout they gave to head and tail players under each endowment (in the main text we show this visualization for endowment [10, 2, 2, 2] in **Fig. 1b**). We used simulated data in which the behavior of virtual human players was unrolled in repeated games of 10 trials. In each game, the payouts were determined by one of 100 mechanisms (each defined by a parameterization of $v$ and $w$ linearly sampled in 10 bins from 0 to 1; see main text and below for more details) under the 5 endowment conditions. This yielded 10 average relative payouts for both head and tail players per mechanism and endowment [20 features] for 500 mechanisms [5 endowment100 instances]. We used multidimensional scaling (MDS) to reduce the dimensionality from 20 to 2 dimensions and plotted the 100 mechanisms for endowment condition [10, 2, 2, 2] all together (**Fig. 1b**). Other endowment conditions are shown in Fig. S9.

## Overview of the mechanism training method

Our approach to AI mechanism design consisted of three steps: 1) we trained *virtual human players* with supervised learning to imitate human play in the investment game using an existing dataset, which we describe as "pilot" data for the main experiments presented here; 2) we trained a vote-maximizing-mechanism using a deep reinforcement learning (RL) agent interacting with these virtual players; and 3) we evaluated the mechanism by deploying it with new unseen human participants, together with comparison baselines (see above). This last step yielded new data, which could be used to repeat the steps above, and refine both the virtual players and the mechanism. In this section we describe these steps in detail.

# Virtual players - modelling

## Training virtual players using imitation learning

We drew upon data from previous pilot experiments (see "pilot testing", below) to train virtual human players using imitation learning. All data was collected from human participants playing the 10-round investment game (described above) under a variety of different mechanisms and with the same endowment conditions as in our study (but not equally distributed). While generally we included only games in which all players finished the experiment, the training data included data from a handful of pilot games where one or more players dropped out. However, the modelling excluded the responses from the players that had been replaced with randomly responding bots. For more details, see "pilot testing" below.

We used imitation learning in which virtual players are trained to imitate human play. Virtual human players were deep neural networks. Each network was a simulation of a single player, which received information about all players' contributions and receipts on the current trial (just like real human players) and was trained to predict the focal player's contributions on the next trial. As they were equipped with recurrent memory (long short term-memory units or LSTMs) (37) the networks could potentially learn to use the trial history going back to the start of the game to make this prediction.

The network received the following information as input on each step: each player's endowment (4 inputs); each player's previous contribution (4 inputs); each player's previous contribution relative to endowment (4 inputs); and each player's payout (4 inputs). Payouts, endowments and contributions were divided by 10 to lie in approximately the same range as relative contributions. These inputs were fed to a linear layer with output size of 64 and *tanh* non-linearities, followed by an LSTM with hidden size 16. The LSTM output to a final linear layer of size 11 whose outputs coded the unnormalized log-probabilities of a categorical distribution corresponding to the probability of contributing 0, 1, …, 9, 10 coins. We masked those outputs corresponding to contributions in excess of the endowment allocated to the focal player.

We trained this architecture with back-propagation through time (38) to minimize the cross-entropy between the predictions and the actual contributions, regularized with two additional terms: the entropy of the prediction (with weight 0.1); and the L2 loss on the parameters of the network (with weight 0.00001). The model was implemented in TensorFlow 1 and optimized the architecture using Adam (39) with learning rate 0.0004 and parameters beta1 0.9, beta2 0.999, and epsilon 1e-08. We trained the model by performing 30,000 updates with mini batches of size 512. Training took less than 6 hours without use of accelerators.

The virtual human player networks were evaluated on a separate hold-out dataset consisting of the contributions of a new group of human players (n = 384) that resembled as closely as possible the final conditions under which we expected to evaluate the *HCRM*. We swept over network hyperparameters (including layer widths, number of LSTM units, learning rate, and regularization types) to minimize validation loss.

## Voting model

We learned in piloting that human players' votes are most strongly predicted by the relative payouts $(y_i/e_i)$ they receive under one mechanism or another. We thus used this variable as the basis for virtual player voting. Each virtual player's probability of voting for a mechanism A rather than the rival mechanism B was $p(A) = \Phi[rpay^A - rpay^B]$ where $rpay^M$ is the sum of relative payouts obtained under mechanism $M$ and $\Phi[\cdot]$ is a logistic function with slope $s$. We set $s$ to be 1.4 but similar results were obtained (in terms of mechanism policy, see below) under a wide range of values (see below).

## Rational players without human data

We also built *rational players* that learned with gradient descent to maximize their return within a single block (rather than learning to imitate human play). We assume that these players have full information about the environment and exact knowledge of the redistribution mechanism at each point in training. Each *rational player* was also characterized by two free parameters: a learning rate, sampled from a Gamma distribution (k = 3, $\theta$ = 1); and a *initial generosity*, sampled from a Normal distribution ($\mu$ = 0, $\sigma$ = 1). These parameters were chosen both to match the previous literature and to facilitate convergence near the end of the block (10 rounds). We also trained mechanisms based on different parameters (Gamma $\theta \in \{1,3\}$, Normal $\sigma \in \{1,3\}$) and performed comparisons between the rational player parameters that the mechanism was trained or evaluated on. Overall, we found that the mechanisms trained with the selected parameters generalized numerically better to other rational player parameters (see supplementary table ST2).

Each round, the rational player performed gradient ascent to maximize their immediate reward with respect to their generosity, assuming that other players' contributions were constant (i.e. independent gradient ascent).

The equations describing the updates of the *rational players* are as follows:

$$\alpha_i \sim \Gamma(k = 3, \theta = 1)$$
$$g_i^{\ 1} \sim N(\mu = 0, \sigma = 1)$$
$$c_i^{\ t} = e_i/(1 + exp(-g_i^{\ t}))$$
$$g_i^{\ t+1} = g_i^{\ t} + \frac{\partial}{\partial g_i^{\ t}}[y_i^{\ t} + e_i - c_i^{\ t}]$$

where $i$ is an index corresponding to each player, $g_i^{\ t}$ is an intermediate variable that we call "generosity", and $\alpha_i$ is the learning rate at which contributions were updated. Note that the derivative in the last equation depends on everyone's contributions, the scaling factor $r$ of the investment game and the mechanism that redistributes the payouts. In practice, we automatized the computation of this derivative with back-propagation. So as to not disadvantage them unduly, we allowed the rational players to vote according to the same human-derived policy as the virtual human players that were used to train HCRM.

## Mechanism designer - training

### Problem definition and set-up

We call the neural network used to design the mechanism the *mechanism designer* and use the term human-compatible mechanism (*HCRM*) to refer to the mechanism it designs, which was obtained only after training has converged. It used reinforcement learning (RL) to learn a function that mapped observations (game states generated through interaction with virtual players) onto *redistribution weights* (a variable that determines which player gets what fraction of the project fund). We chose a Graph Network-based architecture (40) that is equivariant to permutation in the ordering of participants and trained it on simulated 10-round investment games with the goal of maximizing the cumulative voting probabilities of the virtual players against a carefully chosen *alternative mechanism* (see below). After training for 10,000 steps, the network parameters were frozen, and the function was exported in a way that allowed ready implementation in a human testing setting.

### Network architecture

Inputs to the network were the endowments, contributions, and relative contributions (i.e. contribution-endowment ratios) for the current round, for all four participants (12 inputs per round). The network's output was 4 numbers that were passed through a softmax function (i.e. so they are positive and sum to 1) to generate the redistribution weights for each player. When we state that the network has no memory, we mean that (1) it does not receive historical information about contributions or payouts as inputs; and (2) it does not have recurrence, i.e. network states are not passed between timesteps. Note, however, that the mechanism could infer implicitly the number of rounds if human/virtual player policies vary their contributions between different timepoints within the block.

We organized the network's observations in a fully connected directed graph $(u, V, E)$ where each player was represented as a vertex $v_k \in V$. Directed edges $e_{s,r}$ connecting $v_s$ and $v_r$ had empty initial attributes, and the input global attribute vector $u$ was empty. Computations in Graph Networks start by updating the edge attributes, followed by the node attributes, and finally global attributes. In particular, directed edge attributes are updated with a function $\varphi_e$ of the input edge attribute, the sender and receiver vertex attributes, and the global attributes vector: $e_{s,r}' = \varphi_e(e_{s,r}, v_s, v_r, u)$; vertex attributes are updated as a function $\varphi_v$ of the input vertex attributes, the sum of all updated edge attributes that connect into $v_r$, and the global attributes vector: $v_r' = \varphi_v(\Sigma_s e_{s,r}', v_r, u)$; finally, the global attributes vector is updated with a function of the input global attributes, and the sum of all updated edges and vertices: $u' = \varphi_u(\Sigma_{s,r} e_{s,r}', \Sigma_k v_k', u)$. We note that the same functions $\varphi_e$, $\varphi_v$ are used to update all edges and nodes in a graph, and that both the input and output of Graph Networks are directed graphs, so these modules can be used in sequence.

The mechanism designer's policy network architecture consisted of two Graph Networks (GNs) that processed the observation in sequence. In the first GN, we implemented all of $\varphi_e$, $\varphi_v$ and $\varphi_u$ as distinct linear layers with 32 output units and tanh activation functions. In the second GN we implemented $\varphi_e$ as a linear layer with 32 output units and tanh activation function, and $\varphi_v$ as a Linear layer with a single output unit. We normalized the vertex outputs with a soft-max across players, thus obtaining the redistribution weights; $\varphi_u$ was ignored in the second GN.

## Training algorithm

To train the mechanism, we used an estimator drawn from a framework known as Stochastic Computation Graphs (Schulman et al., 2016), which approximates the gradients of the vote-maximization objectives with respect to the network's policy parameters. We trained the mechanism designer iteratively with 10,000 updates using the RMSProp algorithm to optimize the policy, with the following parameters: learning rate 0.0004; epsilon 1e-5; decay: 0.99; and no momentum. On every update, we simulated two batches of 512 games with 10 rounds per game. We divided the batches in groups of 64 episodes with consistent endowments, leading to 8 possible endowments: the head player received 10 coins, and the tail players received 2, 3, 4, 5, 6, 7, 8 or 10 coins (using a broader range of tail player endowments helped avoid overfitting).

On every round the game unfolded as described above for human players, except that the contributions were dictated by the virtual human players. In the first block, the redistribution policy was decided by the mechanism designer under training, and in the other it was played by an alternative planner (which was the winner of the metagame, defined by $w = 1, v = 1$; see section on ideological manifold). We paired episodes from these two batches in order to obtain 2048 votes (512 pairs of episodes, 4 players) given our model of human voting. The objective that the HCRM aimed to maximize was the sum of votes across players, averaged across episodes.

Note that during training of the mechanism, we did not feed in the human data to predict player contributions (i.e. "teacher forcing"). Furthermore, the payouts observed by the virtual players depended on mechanism policies that may lie outside of the human data, thus requiring the virtual players to generalize beyond the training dataset.

Having defined the observations as well as the objective that we wished to maximize, we estimated the *policy gradient*, that is the gradient of the objective (the average number of votes) with respect of the policy parameters (the weights of the graph network) by turning to the Stochastic Computation Graphs (SCG) framework (27). We note here that most of the computation in the investment game is differentiable (including the policy as implemented by the HCRM), with the virtual human players policies, whose action space is discrete, being the only exception. The SCG framework generalizes the policy gradient theorem (41) and allowed us to obtain a low-variance estimator of the policy gradient by auto-differentiating through the environment and mechanism policy, whilst compensating for the non-differentiable operations (the discrete contributions of the players). The surrogate objective for the policy gradient was as follows:

$$S = J + \perp (J) \times \Sigma_i \Sigma_{t=2}^{10} log p(\perp (c_i{}^t))$$

where $S$ is the surrogate objective, $J$ is the objective we wish to maximize per episode (the expected number of votes) and $\perp$ is the stop-gradient operation. Note that for the second term, the gradient can only flow through the parameterization of the log-probability of the player's policy. Note also that the contributions of the first round are removed from the equation, since they do not depend on the mechanism's parameters. In practice, additionally, we chose to de-mean $J$ within a batch because this is known to reduce the variance of the gradient estimator.

## Metagame (or round robin tournament)

We trained the mechanism designer to maximize the votes it is expected to receive in an election against an alternative mechanism. We call this a "meta-game". We selected the alternative mechanism by simulating round-robin elections between candidate mechanisms defined by linearly sampled values of $v$ and $w$ (see ideological manifold section for details of how these parameters are used to define mechanisms). We used virtual human players and selected the mechanism that won the highest number of votes against all opponents over games of 10 rounds (we found no Condorcet cycles in our round-robin elections) to function as the alternative mechanism during training. The expected number of votes received by each mechanism was estimated from a total of 4096 independent blocks. The winning mechanism was defined by $v = 1$ and $w = 1$, which corresponds to liberal egalitarian, and which we subsequently implemented as one of our baselines. Results of the round-robin tournament are shown in Table S2.

## Pilot testing

The data used for training used a variety of data from different experiments (n > 4000, not including data from Exp1-3), all similar in form to the experiments presented above. The games covered a range of different endowment settings and mechanisms, including earlier incarnations of the *HCRM*. In a subset of the training data participants played 4 blocks with referees and voted twice. Overall, the training set contains 4809 blocks, of which 966 were played under equal endowments, 326 under [10, 4, 4, 4], 322 under [10, 2, 2, 2], 1271 under [10, 5, 4, 3], 324 under [10, 8, 5, 3], and 300 under [10, 9, 8, 3]. For the remaining 1300 pilot participants, the endowment was randomly selected between 1 and 10. Note that within any given block, we only model data from participants who did not time out on any trials (i.e. for whom there are 10 contributions in that block). However, the training data (unlike the validation dataset or Exp. 1-3) contains blocks in which participants dropped out of the game and were replaced with a randomly responding bot (2491 blocks contained all four players, the other blocks contained one or more player who dropped out).

The additional validation dataset contained 288 blocks covering the same conditions as Experiments 2-4. Note that this way the validation set tested a degree of generalization to the [10, 6, 6, 6] and [10, 8, 8, 8] conditions which were not included in the training data.

## Human referee experiments

To test whether humans could design a similarly successful mechanism, we implemented a version of the game involving four players and an additional participant who takes the role of the referee and redistributes the funds on every round. In what follows we use the term "player" to refer to the participants who receive the endowment on each tround, and "referee" to the participant who redistributes the public fund back to the players.

**Endowment condition**

In this new experiment, we focussed on the [10, 4, 4, 4] endowment setting. This choice was made for several reasons. Firstly, [10, 4, 4, 4] and [10, 2, 2, 2] were the endowment conditions that (on average against all other competing mechanisms) the agent found most challenging. However, [10, 2, 2, 2] is a less interesting case because the tail player has a highly restricted action space (can only contribute 0, 1 or 2 coins), and simulations with virtual citizens show that liberal egalitarian is optimal in this condition. By contrast, in [10, 4, 4, 4] the agent has to carefully partition funds between the head player (who has the most power to fund the public purse) and the tail players (who collectively have the most votes). We thus thought that it presented an interesting challenge for our agent to attempt to beat human referees in this endowment condition.

**Task (player perspective)**

From the perspective of the four players, the new game was identical to the setting with two algorithmic referees. After playing a game under the egalitarian baseline, the players played two games, one with the human referee and one with HCRM. As before, on each of 10 rounds, they made contributions to the private or public funds, and received back any redistributed funds. They then voted, and experienced 4 trials of the referee (human or HCRM) that won the vote.

Unlike HCRM, human referees do not issue decisions instantly, and so we additionally introduced a mechanism to equate the timing of redistribution decisions between HCRM and the human referee, to avoid the possibility that an automated mechanism is preferred because it is faster (thus increasing participants payouts per hour). To rule out this possibility we sampled HCRM reaction times from reaction times recorded in a pilot human sample. In practice, on average these sampled times were *slower* than those issued by human participants in the main experiment (mean (standard deviation): 40.36 (20.74) vs. 33.14 (14.1) seconds; $t_{60} = 2.2$, $p < 0.02$). Whilst we think it likely that participants would have preferred the faster referee, to test whether voting decisions depended on timing, we correlated RTs of human referees to their percentage votes obtained (out of all players that completed the game), finding no relationship (Pearson's $r_{60} = 0.15$, $p = 0.24$). In combination, these two factors make it unlikely that the results we obtained were driven by any differences in RTs.

**Referee Recruitment**

We wanted to ensure that participants who take the role of the referee were familiar with the task and interface, so that they understood the incentives and dynamics of the game, allowing them to implement the redistribution policy that they think will maximise votes. Thus, to be eligible to participate as referee, participants must have taken part in a previous version of the study (over the past ~12 months) in the role of a citizen.

**Referee Training**

To introduce referees to the basic game mechanics and their role, they first completed a detailed tutorial with instructions on the game worked, what their role would be, and how to redistribute. They then performed a quiz with questions about what they had just learned. Subsequently, they completed 3 blocks of 10 rounds playing with virtual players. At the end of each block, players were informed how many votes they received from the virtual players, virtual player votes were determined in the same way as when training HCRM (by simulating games involving the human allocations and those from HRCM with and calculating the votes of virtual citizens based on this comparison; note that this process was not visible or known to the human referees). This training procedure that the referees underwent thus precisely mirrors the training procedure of the HRCM, except that human referees were only exposed to 3 training episodes. Additionally, prior to performing the main task, referees were again instructed and then performed the quiz, before completing a shorter (5-trial) version of the training (again with virtual players) to refresh their familiarity with the interface. Overall, the training session was completed by participants in between 30 and 60 minutes. Only participants who completed the training were able to sign up for the actual task that was conducted on the same or next day.

**Quiz**

To verify that referees paid attention and understood the basics of the game, they were asked to answer 14 four-way multiple-choice questions about game scenarios. Each question involved plots reporting some aspect of a round (e.g. the distribution of public contributions over players as a bar plot). They were required to answer questions such as: 'How much did player 4 receive from the referee?' or 'How many coins were available to redistribute for the referee?'. After each question participants received feedback on whether they got the question correct or not. In the event of an error, they were informed of the right answer and verbally oriented to the relevant portion of the plot. For example: 'Player 4 received a total of 6.0 coins! You can see this in the most bottom panel. It displays the final earnings. Participants performed the quiz twice with questions in random order, once after instructions for the training and once after instructions for the refresher task. By the time of the refresher, referees scores 94.6% on average (minimum score 71.4%).

An example quiz screen is shown below.

## How many coins were available to redistribute for the referee?

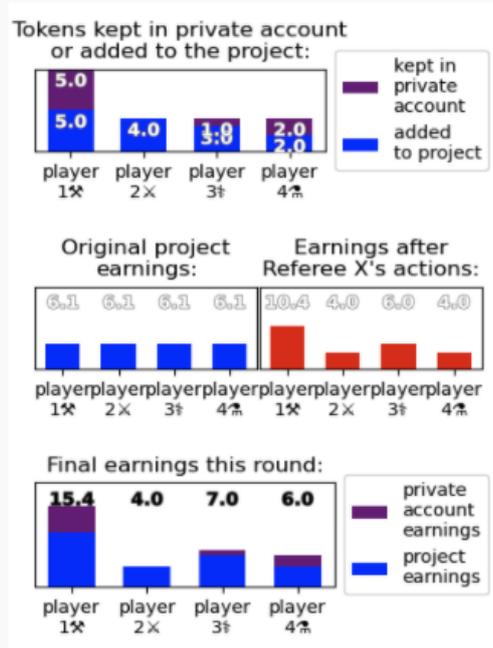

○ 20.4    ○ 20.8    ○ 24.4    ◉ 24.8

CONTINUE TO TUTORIAL ROUND

## Feedback: Wrong! A total 24.4 coins were able distributed back! You can see this in the middle panel.

**Interface**

During the experiment, human referees saw the following interface. It displays how much each player contributed and allows the referee to control the fraction of the pool each player will get as payout in the current round.

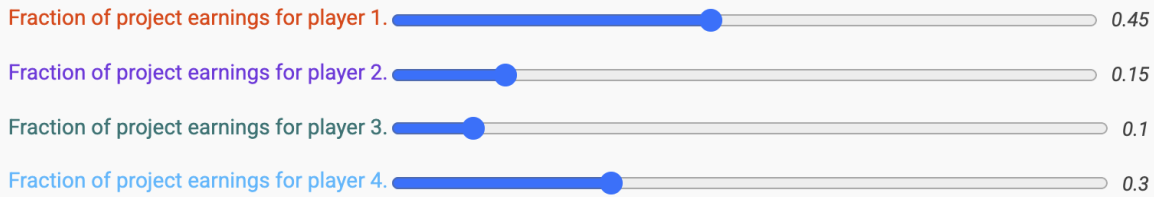

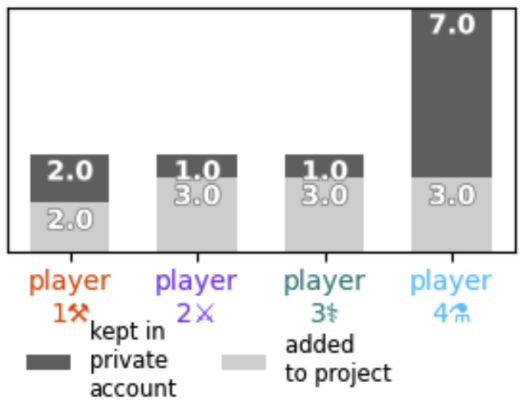
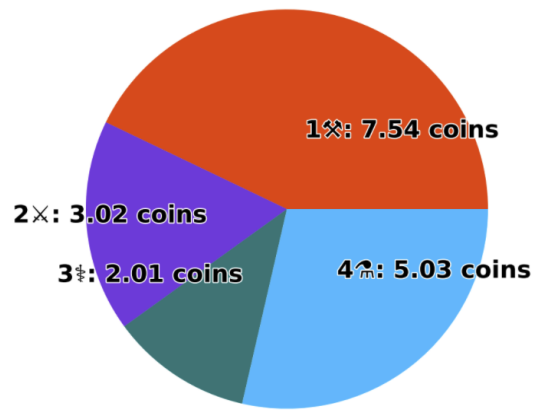

## Data Analysis

We analyzed 61 games in which the human referee completed the task. These included some games in which players dropped out during the experiment (and were replaced as described previously). This resulted in a dataset with 219 votes cast by players (43 games with 4 votes cast, 13 games with 3, 3 games with 2 and 2 games with 1 vote cast). Our primary analysis is of the fraction of votes received by HCRM relative to the human referee. We conduct statistics on the votes of all players regardless of whether their co-players dropped out or not. However, we note that of the 172 votes cast in games where all players finished (n = 43), the players also preferred the HRCM on 62.2% of votes (p < 0.002, Binomial test).

We conducted several analyses aimed at revealing differences between the redistribution policies of the human and HCRM. Firstly, we plotted the distributions of fractional payouts to the head player $payout_{head} / (payout_{head} + payout_{tail})$ under the human and algorithmic referee in six equally spaced bins.

An ANOVA revealed that there was an interaction between sextile and referee, driven by a greater tendency for HCRM to make smaller payouts to the head player (Fig. 4B). Secondly, we plotted how the payout to head and tail (average) players in each game depended on their contribution (Fig. 4C). Finally, we constructed empirical "beach plots" for HCRM and human referees. Note that whereas for algorithmic referees we can probe their response to all possible combinations of contributions by the head and tail player, for human referees we are limited to those that occurred in the actual games (and so beach plots may look slightly different to those shown in **Fig. 3A**). For comparability, we calculated both beach plots in the same way, so the HCRM plot differs slightly from that in Fig. 4.

**Supplementary Text.** Potential limitations of *Democratic AI*.

Our work describes a recipe for training an AI system to design a social or economic mechanism for humans. Our success metric is that humans, having experienced the mechanism, will vote for it in a majoritarian election. Our work thus offers a practical way to deal with issues of value alignment in AI research, in that maximizing human preferences (as expressed through votes) is directly set as the agent's goal (it is part of the cost function). However, we acknowledge that there are both practical and conceptual challenges associated with this approach. We use this supplementary section to highlight potential limitations of our approach, and how they might be overcome in future work.

The first and most important point is that our work should *not* be interpreted as advocating for a form of "AI government", whereby autonomous agents make policy decisions without human intervention (32, 33). We propose *Democratic AI* as a research pipeline for discovering social or economic mechanisms that humans may prefer, but it does not imply in any way that final decisions over deployment should be ceded from human to agent. This follows a tradition in the study of technocratic political apparatus that distinguishes between policy development and policy implementation, with the latter remaining in the hands of elected (human) representatives (34).

In fact, our approach deliberately involves human researchers "in the loop" at multiple stages (in addition to human participants, who provide data from which the agent ultimately learns). This includes an end stage that involves interpreting the mechanism designed by the AI, and using standard metrics to study its economic properties (e.g. the extent to which it promotes productivity vs. equality). However, it also includes various stages within the design process. For example, humans are responsible for choosing the space of *rival* mechanisms. An election is a popularity contest among policies or representatives: in our pipeline, humans retain control in part by specifying the competitors (in the current experiment, this is the space of mechanisms that lie on the ideological manifold). In our vision of its deployment, thus, human policy makers remain the final arbiters of whether any new mechanism is likely to be safe, viable, fair or effective, and no final autonomy over governance is ceded to the AI.

A second important point concerns the interpretability of the AI-designed mechanism (9). We deliberately hampered the mechanism designer by denying it activation memory. This means that the mechanism it designed (*HCRM*) can be transparently described in just two dimensions (rather than, say, being a complicated nonlinear function of the choice history of different players). This is a level of complexity that is similar to the human-generated theories of distributive justice that we use as baselines. Encouraging a more interpretable mechanism has at least three advantages. Firstly, it made the agent more transparent to the human players. In fact, humans deemed the agent to be "more transparent and predictable" than the alternative AI-designed mechanism (*rational mechanism*), as well as (perhaps incongruously) than *strict egalitarian*.

Thirdly, the lack of memory has implications for user privacy. Inputs to the agent were designed to be entirely "slot equivariant", meaning that the mechanisms treated each player's input independent of its 'slot' (ie. if a player is Player 1, 2, 3 or 4). The agent's input pertained to the *distribution* of contributions rather than contributions from individuals themselves. Coupled with the lack of memory, this means that the agent is barred from tracking information about a particular player's history of contributions within the game. The agent applies a general standard to everyone as redistribution is not tied to individual players' identities or history. The slot equivariance also prevents the agent to enact any policy a priori disadvantages any player position ie 'never distribute anything to Player 2'.

Finally, questions might be raised by our emphasis on a democratic objective as a method for value alignment. *Democratic AI* inherits from other democratic approaches a tendency to enfranchise the majority at the expense of the minority: the "tyranny of the majority" (35). This is particularly pertinent given the pressing concern that AI might be deployed in way that exacerbate existing patterns of bias, discrimination or unfairness in society (36). In our investment game, we sampled endowment conditions to match plausible real-world income distributions, where the disadvantaged inevitably outnumber the advantaged – and so for the specific question of distributive justice that we address, this problem is less acute. However, we acknowledge that if deployed as a general method, without further innovation, there does exist the possibility that (just like in real-world democratic systems) it could be used in a way that favors the preferences of a majority over a minority group. We take this risk seriously and note that, just like in the real world, there are no straightforward solutions to this.  However, we would also note (without necessarily endorsing this perspective) the tools that are used in our project at least lend themselves to the implementation of various forms of correction that might be desirable to prevent disenfranchisement of minorities.  For example, it would (in theory) be possible to augment the cost function in a way that hybrid policies, for example in which a democratic objective was mixed with hard or soft constraints that encouraged fairness or protected specific groups. We are not advocating for or against such an approach, but merely pointing out that our research pipeline does not preclude it.

**Fig. S1**. Comparison of virtual human players and human behavior

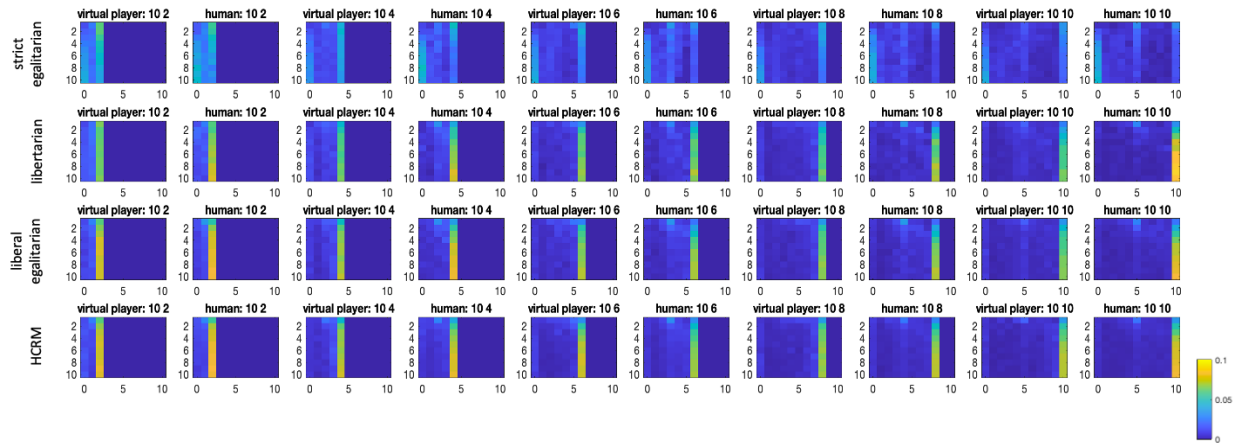

**Fig. S1.** Each plot shows the normalized distribution of coins contributed (x-axis of each plot, from 0-10) over trials (y-axis of each plot, from 1-10), under different endowment conditions and mechanisms (rows top to bottom: strict egalitarian, libertarian, liberal egalitarian, HCRM). Note the similarity between "virtual player" and "human" plots for the same endowment condition / mechanism.

## Fig. S2. Predictors of voting

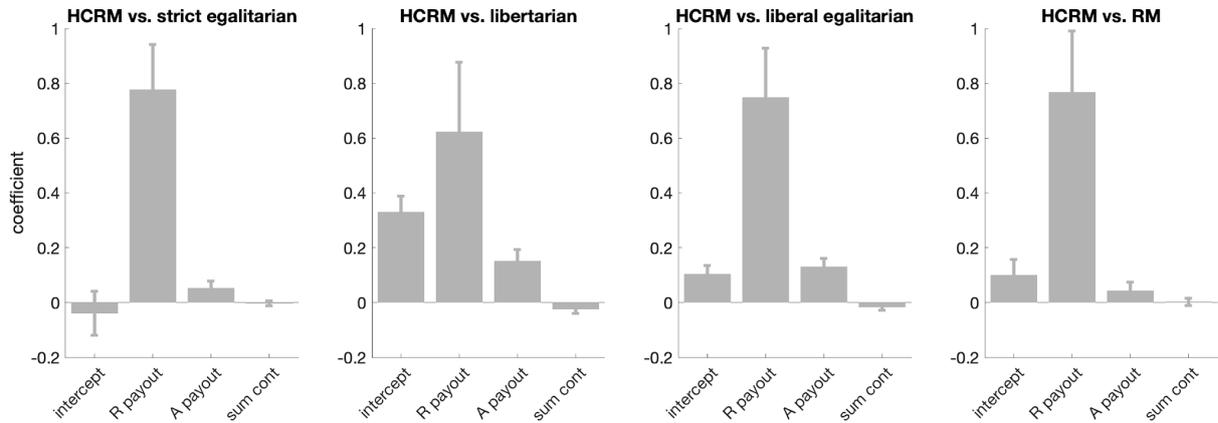

**Fig. S2.** Each panel shows the beta coefficients for a logistic regression predicting votes for the agent vs. a rival baseline (see panel titles) as a function of 4 competitive predictors. "R payout" is the sum of payouts relative to endowment; "A payout" is the sum of absolute payouts; and "sum cont" is total sum of contributions. In all 4 experiments, relative payout was the strongest predictor, and it was significant in all cases. Full statistics are provided in the table below:

| **rival** | Intercept t statistic | Intercept p-value | R payout t statistic | R payout p-value | A payout t statistic | A payout p-value | sum cont t statistic | sum cont p-value |
|---|---|---|---|---|---|---|---|---|
| strict egalitarian | -0.5 | n.s. | 4.7 | < 0.001 | 1.93 | n.s. | -0.3 | n.s. |
| libertarian | 5.57 | < 0.001 | 2.45 | < 0.02 | 3.72 | < 0.001 | -1.7 | n.s. |
| liberal egalitarian | 3.21 | < 0.002 | 4.15 | < 0.001 | 5.31 | < 0.001 | -1.5 | n.s. |
| rational mechanism | 1.81 | n.s. | 3.41 | < 0.001 | 1.29 | n.s. | 0.16 | n.s. |

As can be seen, *relative payout* is the only variable that consistently predicts votes against all 4 baseline mechanisms. This implies that participants consistently normalized their estimates of how generous the agent was in paying out to them by the endowment they initially received. We note that against libertarian and liberal egalitarian the intercept is also reliable. This means that in these conditions, there were additional (unmodelled) variables that predict voting. We can account for some of this variance by including information from the debriefing questionnaire (see Fig. S6).

## Fig. S3. Relative payouts under each mechanism.

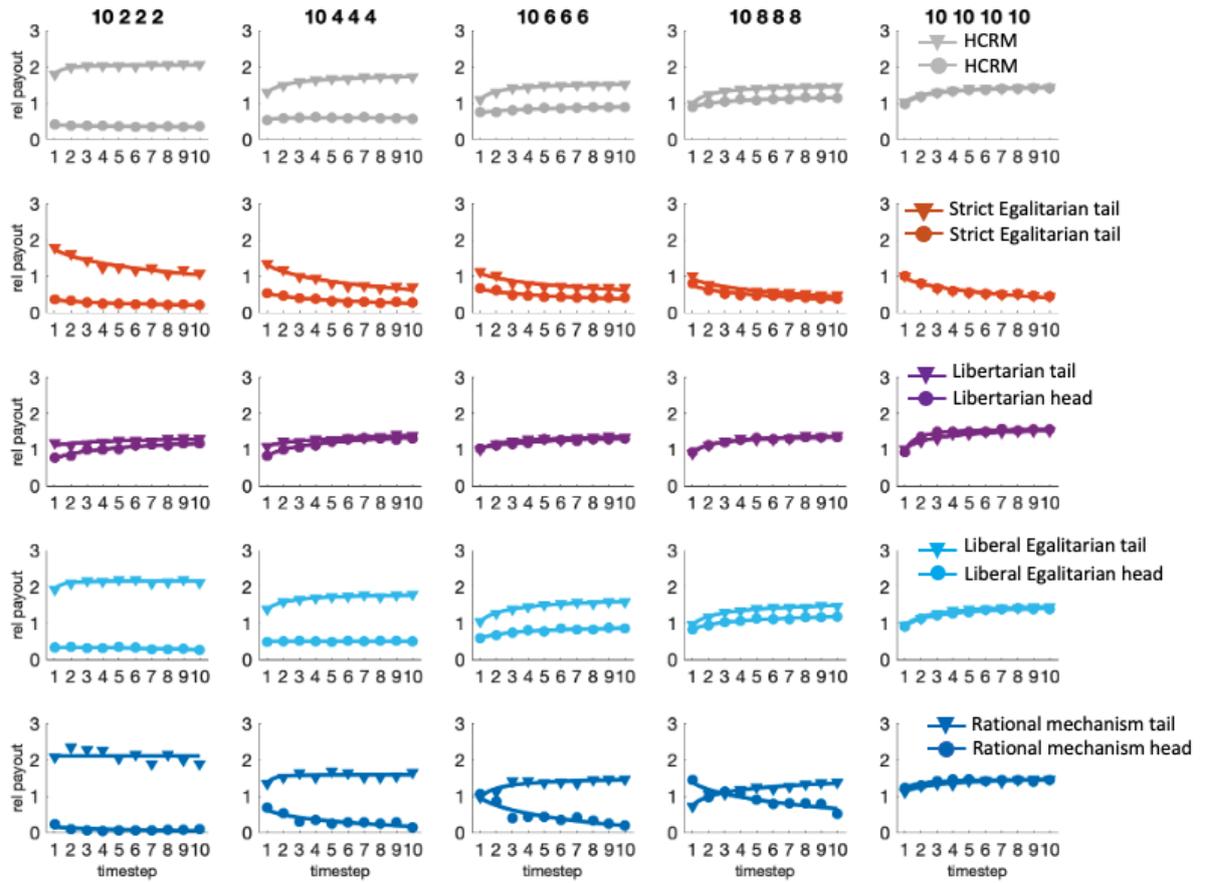

**Fig. S3.** Mean relative payout ($y_i/e_i$) to head player (circles) and tail players (triangles) over each round within the game (x-axis) for the 5 different endowment conditions (columns). Each row is a mechanism. Data are taken from blocks 2-3. The top row is the average of payouts made by the agent against all rivals. Rows 2-5 are payouts made under strict egalitarian, libertarian, liberal egalitarian, and the rational mechanism respectively.

**Fig. S4.** Relative contributions under each mechanism.

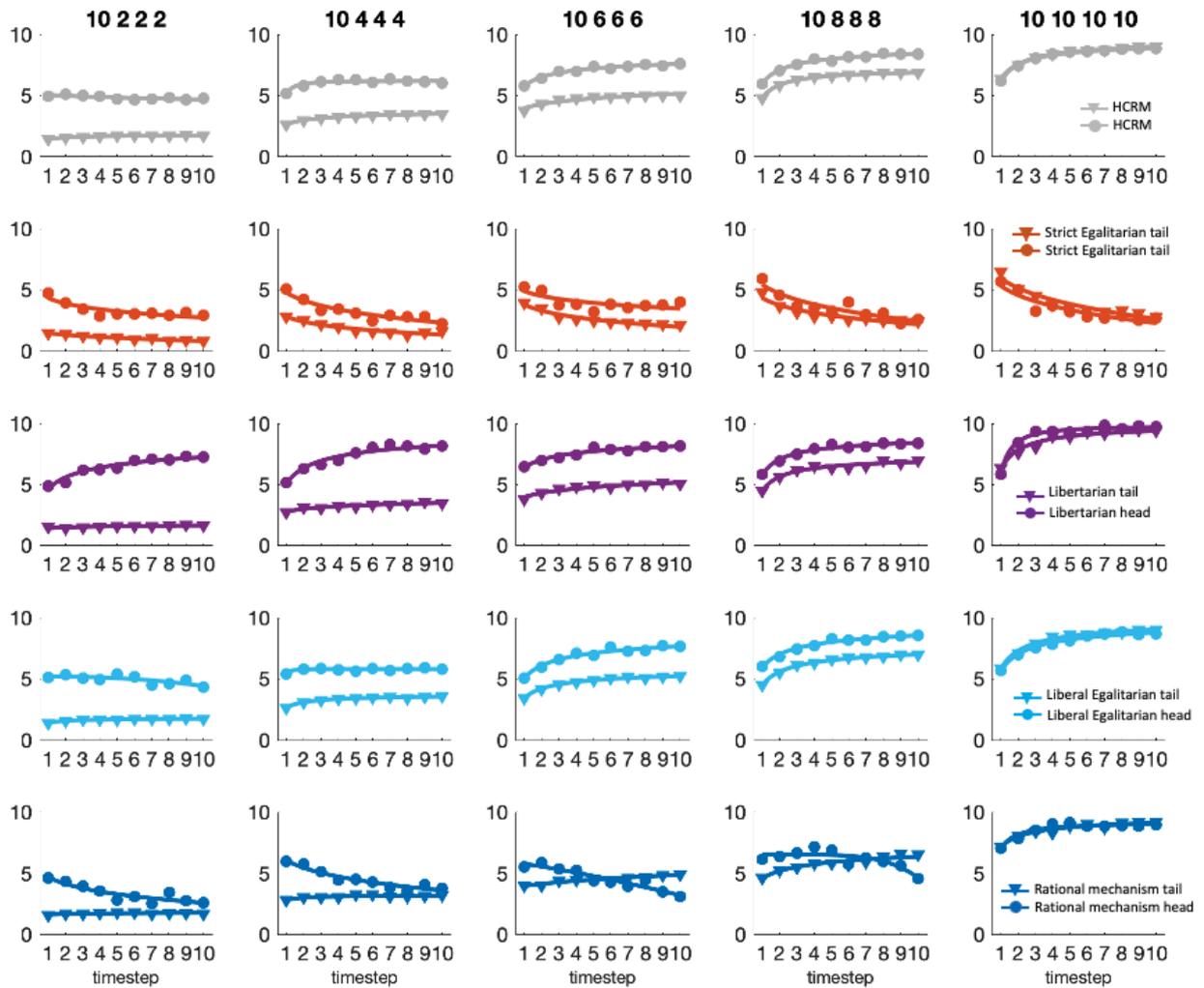

**Fig. S4.** Mean contributions made by head player (circles) and tail players (triangles) over each round within the game (x-axis) for the 5 different endowment conditions (columns). Each row is a mechanism. Data are taken from blocks 2-3. The top row is the average of contributions made by the agent against all rivals. Rows 2-5 are payouts made under strict egalitarian, libertarian, liberal egalitarian, and the rational mechanism respectively.

**Fig. S5.** Distribution of Gini Coefficients for each mechanism

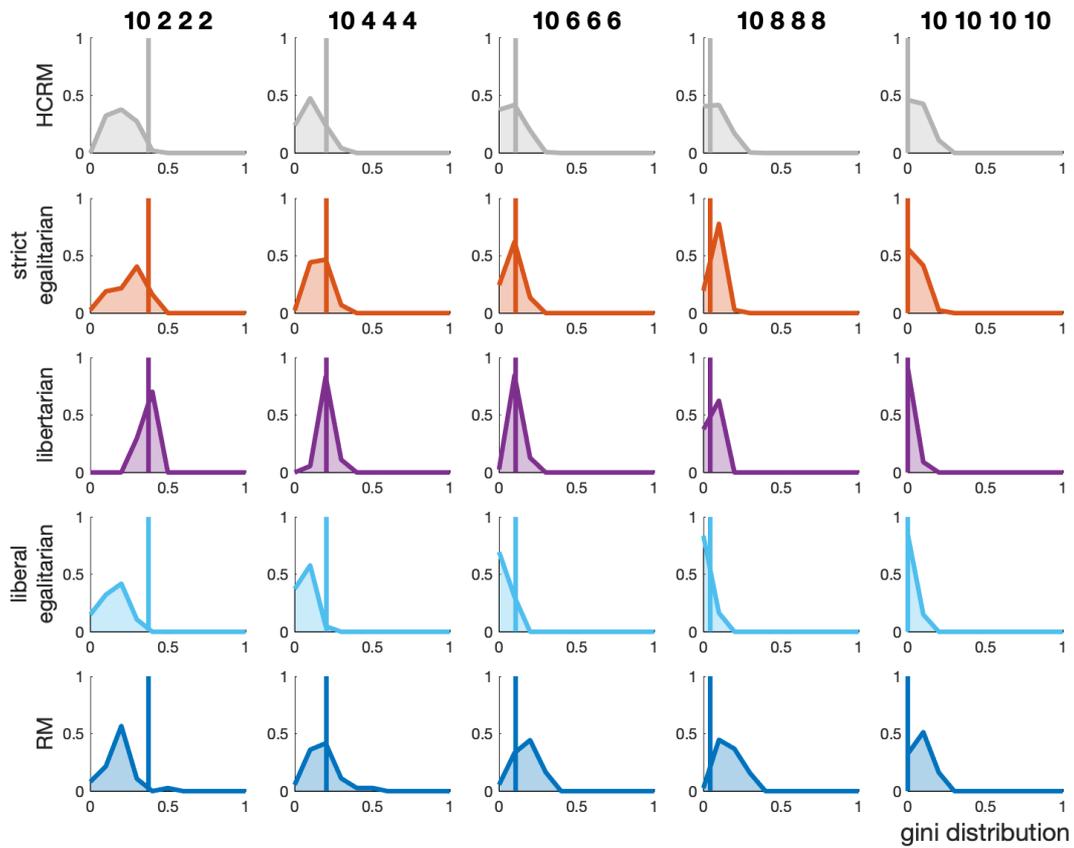

**Fig. S5.** Distributions of Gini coefficients computed from total return (the return to each player is their payouts plus their coins retained in the private fund) to the 4 players at the end of each game. Distributions are normalised so that they sum to one (over ten bins from 0-1). Each column is an endowment condition, and each row is a mechanism. The line shows the Gini for the initial endowment provided on each trial. Lower Gini coefficients imply greater equality.

**Fig. S6.** Debriefing Survey results

**Debrief results.** After voting, but before playing the final round, players completed a debriefing survey, which helped us understand why humans voted for *HCRM*. The survey consisted of 7 questions, to which they were asked to answer "A" or "B".

1. Which referee will lead to everyone doing better?
2. Which referee will lead to you individually doing better?
3. Which will lead to the majority of players doing better?
4. Which referee will be fairer?
5. Which referee will be best at fostering collaboration?
6. Which referee will be more permissive or lenient?
7. Which referee will be more transparent or predictable?

Full percentage votes for *HCRM* for each question (in each endowment condition and against each rival mechanism) are shown below (pink plots). However, we start by noting that positive response to all questions except q.6 predicted voting (all t-values > 4, all p-values < 0.001). Question 2 was by far the strongest predictor (t ~ 26, p ~ 0) confirming that people's voting was mostly, but not exclusively, self-interested.

In the plots below, the format is identical to **Fig. 2** in the main text.

### 'Which referee will lead to everyone doing better?'

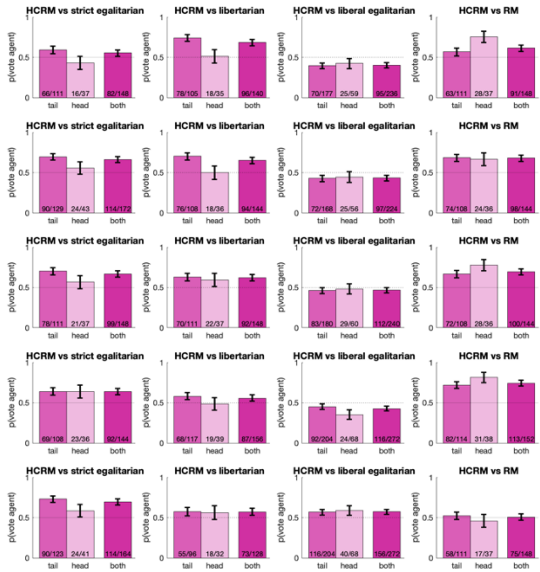

### Which referee will lead to you individually doing better?

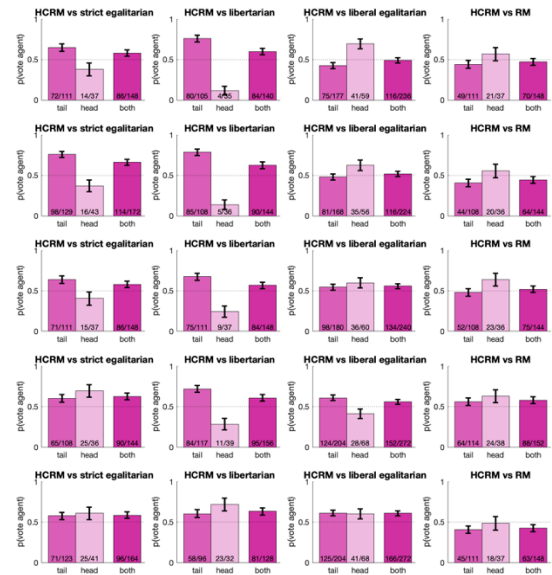

### Which will lead to the majority of players doing better?

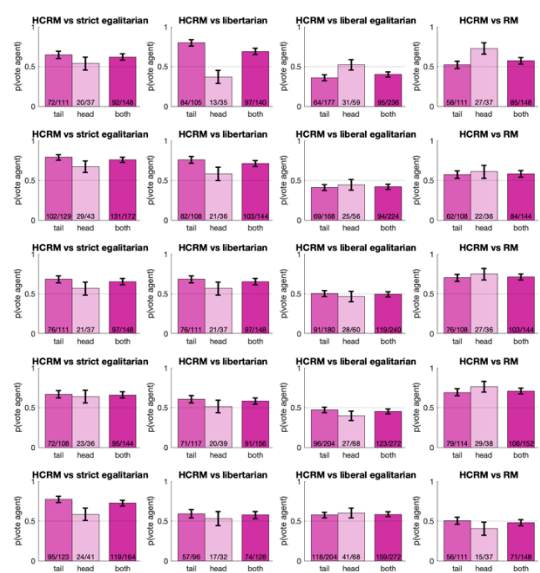

### Which referee will be fairer?

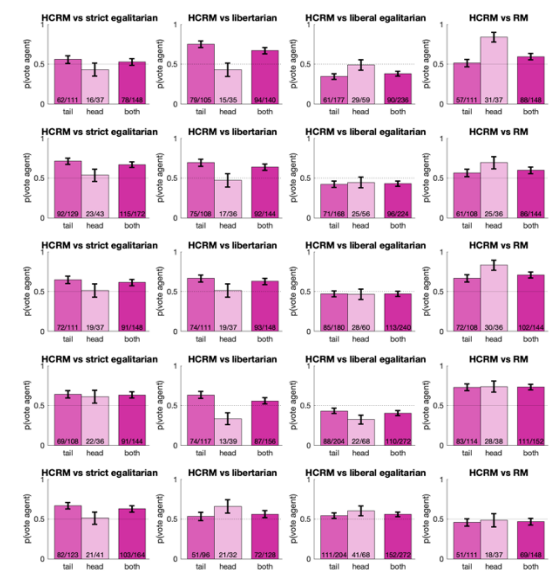

Which referee will be best at fostering collaboration?

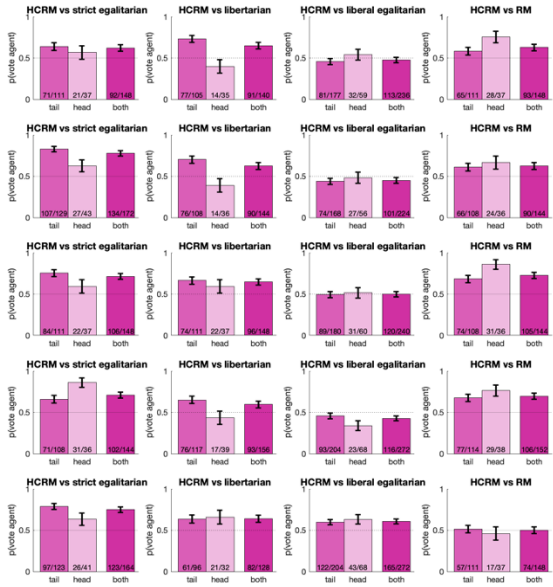

Which referee will be more permissive or lenient?

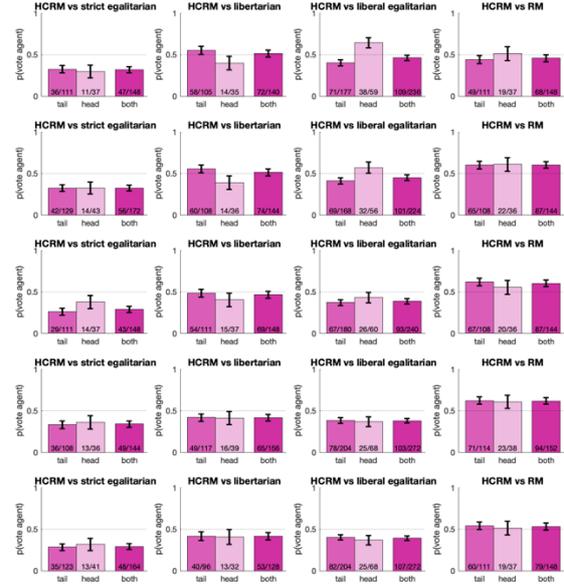

Which referee will be more transparent or predictable?

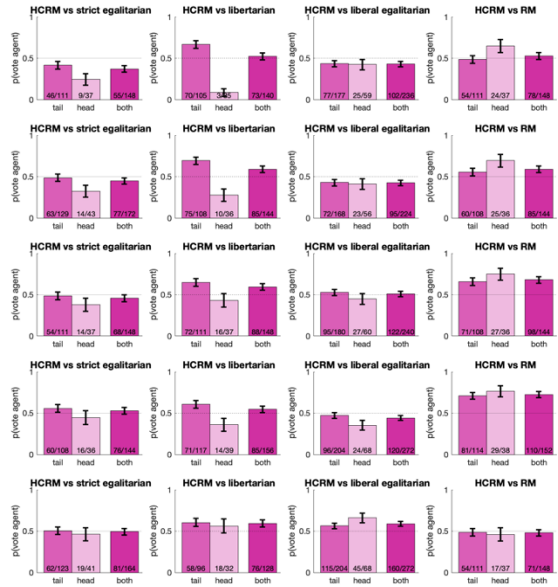

**Fig. S7.** Contribution distribution for each endowment and timepoint

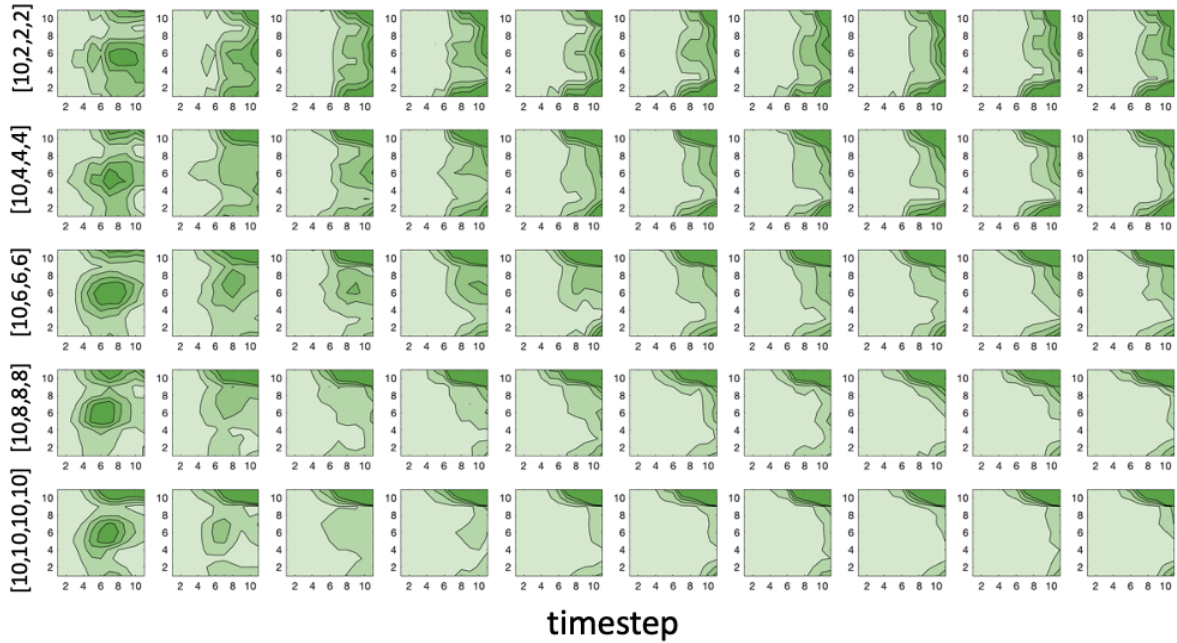

timestep

**Fig. S7.** Relative contribution distributions from the head player (y-axis) and tail player (x-axis), for each endowment condition (row) and round (or "timestep") within the game, averaged over all mechanisms. Darker green shading implies greater density, i.e. a higher frequency of contributions. For example, the approximately trimodal contribution from the head player on round 1 occurs because most players contribute none, half, or all their endowment (0, 5 or 10 coins) on the first round.

**Fig. S8.** Beach plots for return to tail / head player

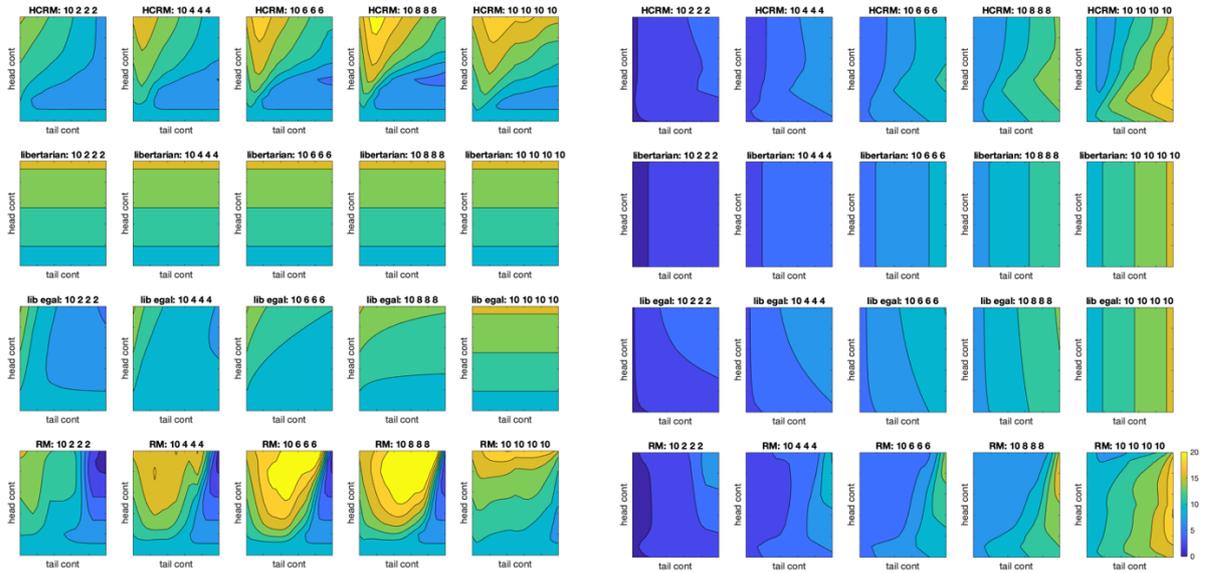

**Fig. S8.** Alternative beach plots describing each mechanism in terms of the return (i.e. payout plus residual endowment) for tail player (left 5 x 5 grid) and head player (right 5 x 5 grid). Units are coins (0-20). Each subplot in the 5 x 5 grid shows return to the tail (left plots) or head (right plots) players as a function of the contributions off the head player (y-axis of each subplot) and tail player (x-axis of each subplot). For example, straight lies horizontal (vertical) for libertarian policy indicate that contributions only depend on tail (head) player. Note that the HCRM creates a strong incentive for the head and tail players to compete to contribute more.

**S9**. The ideological manifold for each endowment condition.

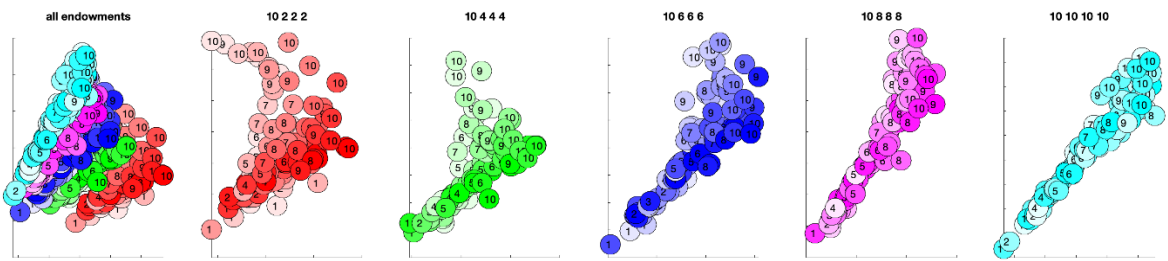

**Fig. S9.** The panel on the left shows the ideological manifold for all 5 endowment conditions together, using the color scheme from panels 2-6. Each panel 2-6 shows the ideological manifold for a specific endowment condition. Each dot is a mechanism. This mechanism was deployed with virtual players and we recorded and concatenated the average (relative) payout to head and tail player over 10 trials (20 features). This was used to determine the similarity among all 100 mechanisms, sampled from the 10 x 10 space defined by bins of mechanism parameters $v$ and $w$. Panel 2 (second from the left) is shown in **Fig. 1b** of the main text.

**Table S1**. Tables of votes.

| | Votes received by HCRM (human players) against each baseline | | | |
|---|---|---|---|---|
| Endowment condition | Strict Egalitarian | Libertarian | Liberal Egalitarian | Rational Mechanism |
| [10 2 2 2] | 84/148 (56.8%) | 92/148 (62.1%) | 162/372 (43.5%) | 82/148 (55.4%) |
| [10 4 4 4] | 121/172 (70.3%) | 96/144 (66.6%) | 172/332 (51.8%) | 74/144 (51.4%) |
| [10 6 6 6] | 93/148 (62.8%) | 91/156 (58.3%) | 191/328 (58.2%) | 96/144 (66.7%) |
| [10 8 8 8] | 99/144 (68.8%) | 92/160 (57.5%) | 186/364 (51.1%) | 98/152 (64.5%) |
| [10 10 10 10] | 116/164 (70.7%) | 79/132 (59.9%) | 240/372 (64.5%) | 71/148 (47.9%) |

**Table S1a.** Votes received by HCRM against each rival mechanism (real huma data). Raw numbers (x out of n votes, upper row) and percentage (lower row), for each endowment condition.

| | Votes received by HCRM (human players) against each baseline | | |
|---|---|---|---|
| Endowment condition | Strict Egalitarian | Libertarian | Liberal Egalitarian |
| [10 2 2 2] | 60.5% | 60.6% | 51.3% |
| [10 4 4 4] | 61.7% | 54.6% | 51.0% |
| [10 6 6 6] | 64.4% | 53.0% | 51.6% |
| [10 8 8 8] | 64.2% | 52.3% | 51.5% |
| [10 10 10 10] | 67.5% | 52.1% | 51.9% |

**Table S1b.** Percentage votes received by HCRM against each rival mechanism for the virtual players. No counts are included because virtual players can be replicated as many times as required.

## Table S2. Results of the round robin tournament ("meta-game")

| | | | | | | | | | |
|---|---|---|---|---|---|---|---|---|---|
| **(v=0, w=0)** | 0.500 | 0.503 | 0.500 | 0.415 | 0.414 | 0.410 | 0.373 | 0.347 | 0.347 |
| **(v=½, w=0)** | 0.497 | 0.500 | 0.503 | 0.418 | 0.417 | 0.416 | 0.372 | 0.349 | 0.343 |
| **(v=1, w=0)** | 0.500 | 0.497 | 0.500 | 0.417 | 0.415 | 0.413 | 0.374 | 0.349 | 0.346 |
| **(v=0, w=½)** | 0.585 | 0.582 | 0.583 | 0.500 | 0.499 | 0.498 | 0.451 | 0.424 | 0.424 |
| **(v=½, w=½)** | 0.586 | 0.583 | 0.585 | 0.501 | 0.500 | 0.503 | 0.453 | 0.428 | 0.425 |
| **(v=1, w=½)** | 0.590 | 0.584 | 0.587 | 0.502 | 0.497 | 0.500 | 0.452 | 0.431 | 0.421 |
| **(v=0, w=1)** | 0.627 | 0.628 | 0.626 | 0.549 | 0.547 | 0.548 | 0.500 | 0.475 | 0.467 |
| **(v=½, w=1)** | 0.653 | 0.651 | 0.651 | 0.576 | 0.572 | 0.569 | 0.525 | 0.500 | 0.494 |
| **(v=1, w=1)** | 0.653 | 0.657 | 0.654 | 0.576 | 0.575 | 0.579 | 0.533 | 0.506 | 0.500 |

**Table S2** Average number of votes given by virtual players for each mechanism generated from the linear redistribution space. We defined a set of 9 mechanisms by exhaustively combining values of v and w in {0, ½, 1}. We created a 9 x 9 payoff matrix corresponding to the average number of votes achieved by each mechanism by exhaustively combining these 9 mechanisms in pairs. These are shown here as win probabilities. The order of mechanisms is matched for rows and columns (leading to a value of 0.5 in the diagonal). The main finding of the meta-game is that *liberal egalitarian* ($v = 1, w = 1$) is a dominant strategy, meaning that it is the strategy achieving the highest number of votes for each player (e.g. rows) independently of the strategy of the opponent (e.g. columns).